\documentclass[letterpaper,journal]{IEEEtran}

\usepackage{amsmath,amsfonts}
\usepackage{graphicx}
\usepackage{booktabs}
\usepackage{multirow}
\usepackage{array}
\usepackage{makecell}
\usepackage{cite}
\let\citep\cite
\usepackage{textcomp}
\usepackage{url}
\usepackage{placeins}
\usepackage{stfloats}
\usepackage[export]{adjustbox}

\usepackage[colorlinks=true,
            linkcolor=blue,
            urlcolor=blue,
            citecolor=blue]{hyperref}
\usepackage[caption=false,font=footnotesize]{subfig}
\newcolumntype{C}[1]{>{\centering\arraybackslash}p{#1}}
\title{Robot-Aligned 3D Gaussian Splatting for Precise Reaching with Wrist-Camera VLA Policies}

\author{Zuoxu Wang, Xiao Liang$^*$
\thanks{Zuoxu Wang, Xiao Liang are with the Zachry Department of Civil and Environmental Engineering, Texas A\&M University, College Station, TX 77843 USA(e-mail: zuoxu@tamu.edu; xliang@tamu.edu).}
\thanks{$^*$ Corresponding Author.}
\thanks{This work was supported by the USA National Science Foundation under Grant No.2422826. Portions of this research were conducted with the advanced computing resources provided by
Texas A\&M High Performance Research Computing}
}

\begin{document}

\maketitle


\begin{abstract}
Efficient robotic data collection remains a major bottleneck for vision-language-action (VLA) policy learning, especially for tasks requiring accurate end-effector positioning and camera-action alignment. This challenge is particularly severe in electronic-waste disassembly, where robots operate in cluttered assemblies of densely packed, visually similar components. Full digital-twin simulation is difficult because many parts are thin, reflective, or mechanically complex, making accurate object-level meshes and contact models difficult to obtain.
In this work, we focus on pre-contact reaching and propose a 3D Gaussian Splatting (3DGS)-based data generation pipeline for wrist-camera VLA adaptation. Our method reconstructs the real workspace with 3DGS and renders diverse wrist-camera observations from robot-aligned poses. These observations are paired with smooth, structured reaching trajectories to provide robot-aligned visual-action supervision for VLA fine-tuning. We further augment the VLA policy with an instruction-conditioned reach verifier that estimates target proximity and provides reliable termination before physical interaction.
Experiments in motherboard disassembly demonstrate precise VLA-based reaching using wrist-camera observations alone, without an external side-view camera. With $\pi$0.5, 1,500 generated trajectories achieve 80.0\% precise-reaching success across three targets, compared with 58.7\% using 150 teleoperated demonstrations, while requiring less than 10 minutes of active human capture.
Project website: \url{https://001-wang.github.io/PrecisionReach-GS/}

\end{abstract}



\section{Introduction}
\IEEEPARstart{P}{recise} reaching is a fundamental requirement for real-world robotic manipulation. Before a robot can grasp, pull, insert, place, cut, or otherwise interact with an object, it must first move its end-effector to the correct target region \citep{marwan2021comprehensive, 11288083, oguntosin2017design, robin2016multi}. This requirement becomes especially challenging in robotic disassembly, where targets include small components such as screws and memory modules as well as larger but fragile components such as GPUs \citep{tian2026redefining, foo2022challenges}. These components are often densely arranged and surrounded by neighboring parts, leaving little tolerance for reaching errors \citep{liu2026self, shi2025viso}. Once the robot reaches the correct interaction region, subsequent actions can often be handled by task-specific controllers or skill modules, making target-oriented reaching a critical bottleneck between visual perception and successful manipulation \citep{le2025application}.

Vision-language-action (VLA) models provide a promising framework for learning such reaching behavior directly from visual observations and task instructions \citep{zitkovich2023rt,kim2024openvla,sapkota2025vision,intelligence2025pi05}. By mapping camera observations and language commands to robot actions, VLA policies can learn visually grounded motion generation without requiring each target or task to be manually encoded. With sufficiently large, diverse, and well-aligned demonstrations, VLA models have the potential to generate accurate and adaptive reaching actions across different target objects, viewpoints, and instructions. This makes VLA particularly attractive for manipulation tasks where the robot must understand both the semantic meaning of the target and its spatial location in the scene. Importantly, in many manipulation tasks, the pre-contact reaching stage occupies a large portion of the VLA-controlled action sequence, as the robot must progressively move from an initial pose to the target interaction region before executing the final physical skill. Therefore, improving the data efficiency and reliability of reaching behavior can substantially benefit the overall VLA manipulation pipeline.

\begin{figure*}[t]
    \centering
    \includegraphics[width=\linewidth]{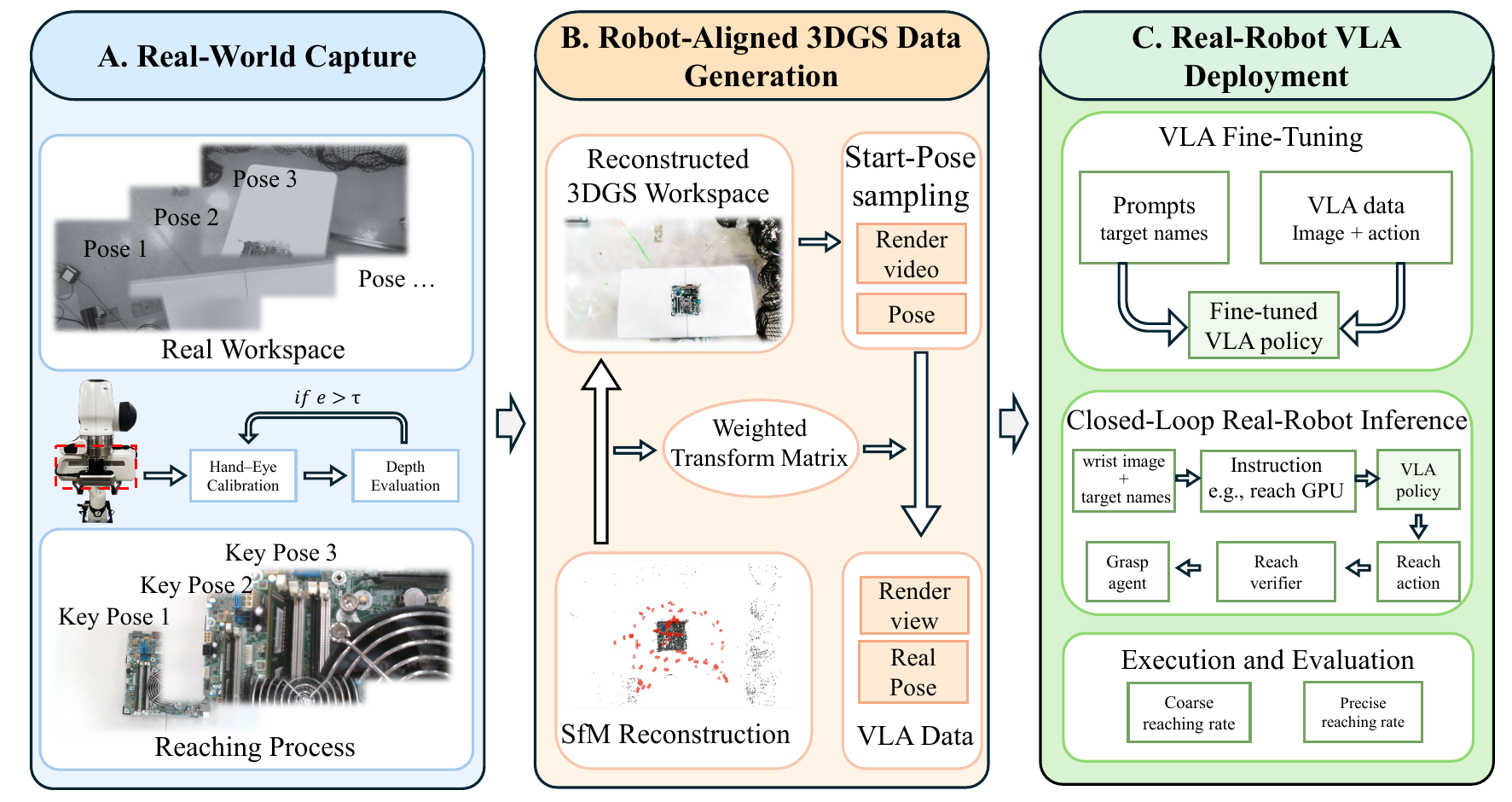}
    \caption{
    Overview of the proposed real-to-sim-to-real pipeline. 
    The system first captures and calibrates the real workspace, then reconstructs and aligns a 3DGS scene for robot-feasible wrist-camera data generation, and finally fine-tunes and deploys a VLA policy for closed-loop precise reaching.
    }
    \label{fig:workflow}
\end{figure*}

However, the effectiveness of VLA models is strongly tied to data scale and data quality \citep{pfaff2025scalable,hofer2021sim2real,mandlekar2023mimicgen,martinez2020unrealrox}.  To learn precise reaching behavior, the policy requires many demonstrations in which the visual observation, language instruction, robot state, and action are consistently aligned. In practice, collecting such data on real robots remains difficult. Most demonstrations are collected through human teleoperation, which is slow, expensive, and difficult to scale \citep{mandlekar2019scaling, zhang2018deep}. This problem becomes more severe when robot access is limited, because data collection occupies the physical robot for long periods and reduces the time available for training, testing, and other experiments. Moreover, human-collected demonstrations often introduce noise that can directly affect VLA learning. Teleoperated trajectories may be jerky, inaccurate, or contain unnecessary corrective motions \citep{richards2025modeling, pmlr-v305-guo25a}. For example, an operator may move the robot toward an approximate region, realize that the target has been missed, and then perform several small adjustments before reaching the desired position. While this behavior is natural during manual operation, it creates ambiguous supervision for the policy: the model may learn both the intended reaching motion and the accidental correction behavior. In addition, delays between camera capture, robot state recording, and action execution can cause visual observations to be paired with slightly mismatched actions. Such camera-action misalignment is especially harmful for precise reaching, where small spatial errors can determine whether the robot reaches the correct component or misses it.

A natural alternative is to generate demonstrations in simulation, but building a full digital-twin simulator for electronic-waste disassembly is also challenging. Motherboards contain many densely packed components with different sizes, materials, and appearances, including small screws and connectors as well as larger but fragile parts such as GPUs, heat sinks, and memory modules. Many of these components are thin, reflective, or mechanically complex, making accurate object-level mesh reconstruction, material modeling, and contact simulation difficult. As a result, full simulation requires substantial manual modeling effort and can still introduce a large sim-to-real gap if the geometry, camera viewpoint, or physical interaction does not match the real workspace.

In addition to data collection and simulation, camera configuration also affects the scalability of VLA training and deployment. Many robot-learning systems use fixed external cameras or multi-view observations, which provide useful global context but require stable placement and calibration \citep{kim2026ego, zhao2026atlasvla}. In contrast, we consider a harder wrist-camera-only setting, where the viewpoint moves with the end-effector and the policy must infer both target location and motion correction from a narrow, continuously changing view. This reduces dependence on fixed side-view cameras, but also makes precise reaching more data-demanding, since the policy needs diverse and well-aligned wrist-camera observations to learn reliable corrections.

In this work, we propose \textit{PrecisionReach-GS}, a lightweight real-to-sim-to-real pipeline for data-efficient VLA fine-tuning in a wrist-camera-only setting. As shown in Fig.~\ref{fig:workflow}, the pipeline first reconstructs the real task environment with 3D Gaussian Splatting (3DGS), preserving the cluttered background, object layout, and visual complexity of the physical workspace. Unlike full digital-twin simulators that require object-level meshes, collision models, and contact dynamics, PrecisionReach-GS focuses on the pre-contact reaching stage, where realistic wrist-camera observations and robot-aligned action labels are the main requirements. The system samples feasible starting poses, renders diverse wrist-camera observations, generates smooth reaching trajectories toward target components, and organizes the resulting observation--action pairs for VLA fine-tuning. During deployment, the fine-tuned VLA policy predicts reaching actions from wrist-camera observations, while a lightweight verifier determines whether the robot has reached the target interaction region. As a result, the user only specifies the target task and desired data scale, and the pipeline produces structured wrist-camera training data without large-scale teleoperation or a complete physical digital twin.

The main contributions of this work are summarized as follows:
\begin{itemize}
\item We propose \textit{PrecisionReach-GS}, a lightweight real-to-sim-to-real framework that uses 3DGS as a photorealistic workspace representation for VLA adaptation, avoiding object-level meshes, contact models, and full physics-based digital twins.

\item We develop a robot-aligned visual-action generation method that maps the reconstructed scene into the robot frame, samples kinematically feasible starting poses, and generates synchronized wrist-camera observations, robot states, and smooth reaching actions with minimal human intervention.

\item We introduce an instruction-conditioned reach verifier that complements the VLA policy with instruction-conditioned reach assessment, enabling more reliable stopping under the distinct geometric precision requirements of different manipulation targets.

\item We experimentally validate the proposed pipeline in a robotic motherboard disassembly scenario, demonstrating that wrist-camera-only VLA fine-tuning, together with reach verification, can achieve precise target reaching and support successful downstream manipulation skills.

\end{itemize}

\section{Related work}
\label{sec:related work}

\textbf{Robot data collection for VLA adaptation.}
Vision-language-action (VLA) models have recently emerged as a promising paradigm for instruction-conditioned robot control, mapping visual observations, language instructions, and robot states directly to actions. Representative models such as RT-2, OpenVLA, OpenVLA-OFT, Octo, $\pi_0$, and $\pi_{0.5}$ show that large vision-language backbones can be adapted for visuomotor control through action tokenization, continuous action regression, action chunking, or flow-based action generation~\citep{zitkovich2023rt,kim2024openvla,kim2025fine,team2024octo,black2024pi0,intelligence2025pi05}. These models are attractive for robotic disassembly because the robot must interpret component-level semantic targets while producing continuous motion toward the correct interaction region.
However, adapting these models to precise, task-specific manipulation remains limited by the availability of high-quality robot action data. Existing robot data sources mainly rely on large-scale robot logs or human teleoperation~\citep{o2024open,walke2023bridgedatav2,fang2023rh20t}. Robot logs can scale with deployment but are often tied to specific embodiments and task distributions, while teleoperation provides flexible demonstrations but remains slow, labor-intensive, and noisy. This bottleneck is especially severe for component-level reaching in robotic disassembly, where small spatial errors can cause downstream manipulation to fail.

\textbf{Real-to-sim-to-real and 3DGS-based data generation for robot learning.}
To reduce real-robot data requirements, recent work has explored real-to-sim-to-real pipelines that reconstruct real scenes and use simulation, rendering, or generated demonstrations for policy learning\citep{fan2025twinaligner}. Early methods reconstruct object-level meshes from real observations and place them into simulators to generate grasp or manipulation data~\citep{wang2023real2sim2real}. More recent systems extend this idea to broader policy learning, including X-SIM, which reconstructs photorealistic simulation from action-less human videos and distills privileged policies into image-conditioned policies~\citep{dan2025xsim}; Real2Render2Real, which generates robot training data from object scans and a single human demonstration~\citep{yu2025real2render2real}; and EmbodieDreamer, which combines physical parameter alignment with photorealistic video generation for reinforcement and imitation learning~\citep{wang2025embodiedreamer}.

3D Gaussian Splatting (3DGS) further supports this direction by enabling photorealistic novel-view rendering from real-world captures~\citep{kerbl20233dgs}. Recent robotic systems use 3DGS for real-to-sim reconstruction, policy learning, simulation, and evaluation. For example, RoboGSim builds a 3DGS-based Real2Sim2Real simulator with mesh-based digital twins and closed-loop policy evaluation~\citep{li2025robogsim}; PolaRiS reconstructs high-fidelity environments for evaluating generalist robot policies, including wrist-camera policies~\citep{jain2025polaris}; and GaussGym integrates 3DGS rendering into vectorized physics simulation for pixel-based locomotion learning~\citep{escontrela2025gaussgym}. Related systems such as SplatSim, Robo-GS, and GSWorld also explore Gaussian-based rendering and simulation for robot policy learning or evaluation~\citep{qureshi2025splatsim,robogs2024,jiang2025gsworld}.

\begin{table}[t]
\centering
\caption{Comparison with representative 3DGS-based robotic data generation methods.}
\label{tab:method_comparison}
\resizebox{\columnwidth}{!}{
\begin{tabular}{lcccc}
\hline
\textbf{Method} &
\textbf{3DGS} &
\textbf{Physics Sim.} &
\textbf{Object Mesh} &
\textbf{GPU Simulator Req.} \\
\hline
SplatSim \citep{qureshi2025splatsim}
& \checkmark & \checkmark & $\times$ & \checkmark \\

RoboGSim \citep{li2025robogsim}
& \checkmark & \checkmark & \checkmark & \checkmark \\

Real2Render2Real \citep{yu2025real2render2real}
& \checkmark & $\times$ & \checkmark & \checkmark \\

\textbf{PrecisionReach-GS}
& \checkmark & $\times$ & $\times$ & $\times$ \\
\hline
\end{tabular}}
\end{table}

These works demonstrate the value of reconstructed or generated visual environments for scalable robot learning. However, most of them target full manipulation, object-motion imitation, policy evaluation, or physics-coupled simulation, and therefore often require mesh extraction, object-level interaction modeling, dynamics simulation, or high-fidelity simulation platforms. Such pipelines introduce higher setup cost and can be challenging to build for complex scenarios.

\textbf{Positioning of our work.}
Compared with prior real-to-sim-to-real and 3DGS-based robot learning systems, \textit{PrecisionReach-GS} is not intended to build a full physical simulator or a general policy-evaluation platform. Instead, it provides a lightweight, lower-setup-cost data generation pipeline for wrist-camera VLA adaptation in pre-contact reaching. Our key distinction is that we use 3DGS as a mesh-free visual data generator aligned with the robot frame, generating feasible visual-action supervision without object-level mesh reconstruction or contact simulation.
The method also differs from conventional visual servoing: it learns
language-conditioned component reaching directly from wrist images instead of
requiring an explicitly detected target feature and a hand-designed image-space
control law.

\section{Method}
\label{sec:Method}

\subsection{Overview}
\label{sec:method_overview}
PrecisionReach-GS consists of three stages, as illustrated in Fig.~\ref{fig:workflow}. 
First, we capture the real workspace with a wrist-mounted camera and record a small set of target-specific key poses, while hand--eye calibration establishes the metric relationship between the robot and camera frames. 
Second, we reconstruct the workspace with 3D Gaussian Splatting and align it to the robot base frame, enabling robot-feasible wrist-view rendering and reaching-trajectory generation. 
Third, the generated wrist-camera observations, target instructions, and action labels are used to fine-tune a VLA policy, which is deployed in closed loop with a reach verifier before transferring control to downstream manipulation. 
This design reduces the need for large-scale teleoperation while preserving the visual complexity of the real workspace.

\subsection{Real-Scene Reconstruction}
\label{sec:real_scene_reconstruction}

The first stage reconstructs the real workspace in a robot-referenced form. We first perform hand--eye calibration between the robot end-effector and the wrist-mounted camera. Although CAD can provide a nominal camera-to-end-effector transform, hand--eye calibration captures physical mounting offsets and remains applicable when a complete CAD-to-optical-frame chain is unavailable. Given the calibrated transformation ${}^{E}\mathbf{T}_{C}$, the camera pose in the robot base frame is
\begin{equation}
{}^{B}\mathbf{T}_{C}^{(i)}
=
{}^{B}\mathbf{T}_{E}^{(i)}
{}^{E}\mathbf{T}_{C},
\end{equation}
where ${}^{B}\mathbf{T}_{E}^{(i)}$ is the end-effector pose of the $i$-th observation.

Because precise pre-contact reaching is sensitive to vertical errors---that is,
errors along the local surface-normal or approach direction---we validate the
calibration with a depth-based plane check. Held-out observations of a planar
reference surface are transformed into the robot base frame, and their
point-to-plane residual is measured as
\begin{equation}
e_i =
\frac{
\left|
\mathbf{n}^{\top}\mathbf{x}_i + d
\right|
}{
\|\mathbf{n}\|_2
},
\end{equation}
where $(\mathbf{n}, d)$ defines the fitted reference plane and $\mathbf{x}_i$ is a newly observed 3D point. This check verifies that the wrist-camera geometry is reliable for vertical reaching, where depth errors may cause missed contact or collision with nearby components.

We then record wrist-camera videos of an unstructured workspace and reconstruct it as a photorealistic 3D Gaussian Splatting scene. The scene preserves cables, clustered components, reflective parts, and cluttered background regions without requiring object-level meshes or contact models. Short reaching videos toward target components are also recorded to obtain key poses near the desired interaction regions.

\begin{figure}[!t]
    \centering
    \subfloat[Mesh rendering.]{\includegraphics[width=0.47\linewidth]{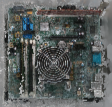}}
    \hfill
    \subfloat[3DGS rendering.]{\includegraphics[width=0.47\linewidth]{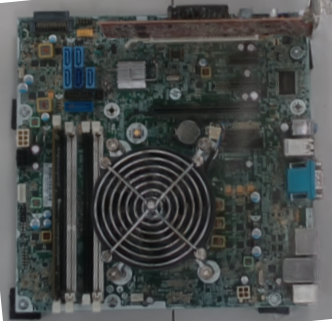}}
    \caption{Rendering representations reconstructed from the same workspace
    capture.  The 3DGS representation better preserves the appearance needed
    for wrist-camera policy observations.}
    \label{fig:mesh_vs_3dgs}
\end{figure}

The choice of 3DGS is motivated by the observation model required for VLA
training rather than by geometry alone.  As illustrated in
Fig.~\ref{fig:mesh_vs_3dgs}, a mesh reconstructed from the same capture exhibits
visible geometry and appearance artifacts even when the motherboard region is
retained.  These artifacts directly alter the wrist-camera input to the policy.
Conventional image augmentation can diversify color or crop statistics, but it
cannot synthesize geometrically consistent observations along a new robot
trajectory.  In contrast, 3DGS preserves the appearance and surrounding context
of the real scene while rendering a continuous sequence of novel views from
robot-referenced poses.

\subsection{Robot-Aligned 3DGS Data Generation}
\label{sec:gs_data_generation}

Given the reconstructed 3DGS workspace, we generate VLA training data through robot-aligned view synthesis. Unlike free-viewpoint rendering, each synthesized wrist-camera view is derived from a robot end-effector pose in the calibrated robot base frame, ensuring that the rendered observations correspond to physically meaningful robot states.

\textbf{Robot-frame scene alignment.}
The 3DGS scene is initially expressed in the SfM coordinate frame, whose scale and orientation are not directly consistent with the robot. We align the reconstruction frame $G$ with the robot base frame $B$ using matched camera centers from SfM and robot kinematics. Specifically, we estimate a weighted similarity transformation by solving
\begin{equation}
\min_{s,\mathbf{R},\mathbf{t}}
\sum_i w_i
\left\|
s\mathbf{R}\mathbf{c}^{G}_i + \mathbf{t}
-
\mathbf{c}^{B}_i
\right\|_2^2 ,
\quad
\mathbf{R}\in SO(3),\ s>0 .
\label{eq:weighted_alignment}
\end{equation}
Here, $s$, $\mathbf{R}$, and $\mathbf{t}$ map the reconstruction frame to the robot frame. The weights $w_i$ emphasize task-relevant anchor frames near target components, improving local alignment accuracy around interaction regions.

\textbf{Feasible start-pose sampling.}
To increase viewpoint and state diversity, we sample candidate starting poses around the reconstructed workspace before generating trajectories toward the target key poses. Because randomly sampled poses may be visually plausible in the 3DGS scene but infeasible for the physical robot, each candidate is filtered using conservative kinematic checks. Let $\Omega_{\mathrm{safe}}$ denote a bounded workspace region around the task area. A candidate pose is accepted only if its position lies within this region, its inverse-kinematics solution has sufficiently small residual error, the resulting joint configuration satisfies the robot joint limits, and the configuration is not close to a kinematic singularity:
\begin{equation}
\begin{aligned}
\mathbf{p} &\in \Omega_{\mathrm{safe}}, &
\epsilon_{\mathrm{IK}}(\mathbf{p}) &< \epsilon_{\max}, \\
\mathbf{q}^{-} &\leq \mathbf{q}(\mathbf{p}) \leq \mathbf{q}^{+}, &
m(\mathbf{q}) &> m_{\min}.
\end{aligned}
\end{equation}
Here, $\epsilon_{\mathrm{IK}}$ denotes the IK residual, $\mathbf{q}^{-}$ and $\mathbf{q}^{+}$ are the robot joint limits, and $m(\mathbf{q})$ measures the local translational manipulability of the arm. We compute
\begin{equation}
m(\mathbf{q}) =
\sqrt{
\det\left(
\mathbf{J}_{p}(\mathbf{q})
\mathbf{J}_{p}(\mathbf{q})^{\top}
\right)
},
\end{equation}
where $\mathbf{J}_{p}$ is the translational part of the robot Jacobian. Low manipulability indicates that the robot is close to a singular configuration, where small Cartesian corrections may require large or unstable joint motions. This filtering step ensures that the generated wrist-camera observations and reaching trajectories correspond to physically executable robot states rather than only visually plausible rendered poses.

\textbf{Trajectory rendering and action labeling.}
For each accepted start pose, we generate a smooth task-space trajectory from
the sampled start pose to a pre-approach pose and then to the target key pose;
we do not use a sampling-based collision-aware planner such as RRT.  Target
poses are recorded during reconstruction and serve both as interaction
keyframes and as high-weight anchors in the scene-to-robot alignment.  Within
each segment, cubic smoothstep timing,
\begin{equation}
    h(u)=3u^2-2u^3, \qquad u\in[0,1],
    \label{eq:smoothstep}
\end{equation}
interpolates translation and the corresponding rotation interpolation.  Its
zero endpoint velocity avoids the abrupt segment transitions produced by
linear timing. Each intermediate end-effector pose is converted to a
wrist-camera pose and rendered from the aligned 3DGS scene. The same generated
trajectory is converted to the action representation required by each VLA
policy. For joint-space policies such as the OpenPI-based $\pi_{0.5}$ policy,
we use inverse kinematics and define the action as
\begin{equation}
\mathbf{q}_{t} = \mathrm{IK}({}^{B}\mathbf{T}_{E,t}),
\quad
\mathbf{a}^{q}_{t} = \mathbf{q}_{t+1} - \mathbf{q}_{t}.
\end{equation}
For pose-space policies such as OpenVLA-OFT, we use the relative end-effector motion:
\begin{equation}
\mathbf{a}^{x}_{t} =
\rho\left(({}^{B}\mathbf{T}_{E,t})^{-1}{}^{B}\mathbf{T}_{E,t+1}\right),
\end{equation}
where $\rho(\cdot)$ maps the relative transform to the policy-specific pose-action format.
For throughput, data-generation workers run in parallel and independently load
the same aligned 3DGS scene.  We apply no worker-specific or render-specific
domain randomization; diversity comes from the sampled robot poses and their
view-consistent trajectories.

\subsection{Policy Learning}
\label{sec:policy_deployment}

The generated trajectories are converted into VLA training samples and used to adapt the reaching policy by behavior cloning. Each sample consists of a wrist-camera observation $\mathbf{I}_{t}$, a target instruction $\ell$, the robot state $\mathbf{s}_{t}$, and a reaching action $\mathbf{a}_{t}$. 
The reaching policy is then trained with the behavior-cloning objective
\begin{equation}
\min_{\theta}
\sum_t
\mathcal{L}_{\mathrm{BC}}
\left(
\pi_{\theta}(\mathbf{I}_{t}, \ell, \mathbf{s}_{t}),
\mathbf{a}_{t}
\right).
\end{equation}

\begin{figure*}[t]
    \centering
    \includegraphics[width=\linewidth]{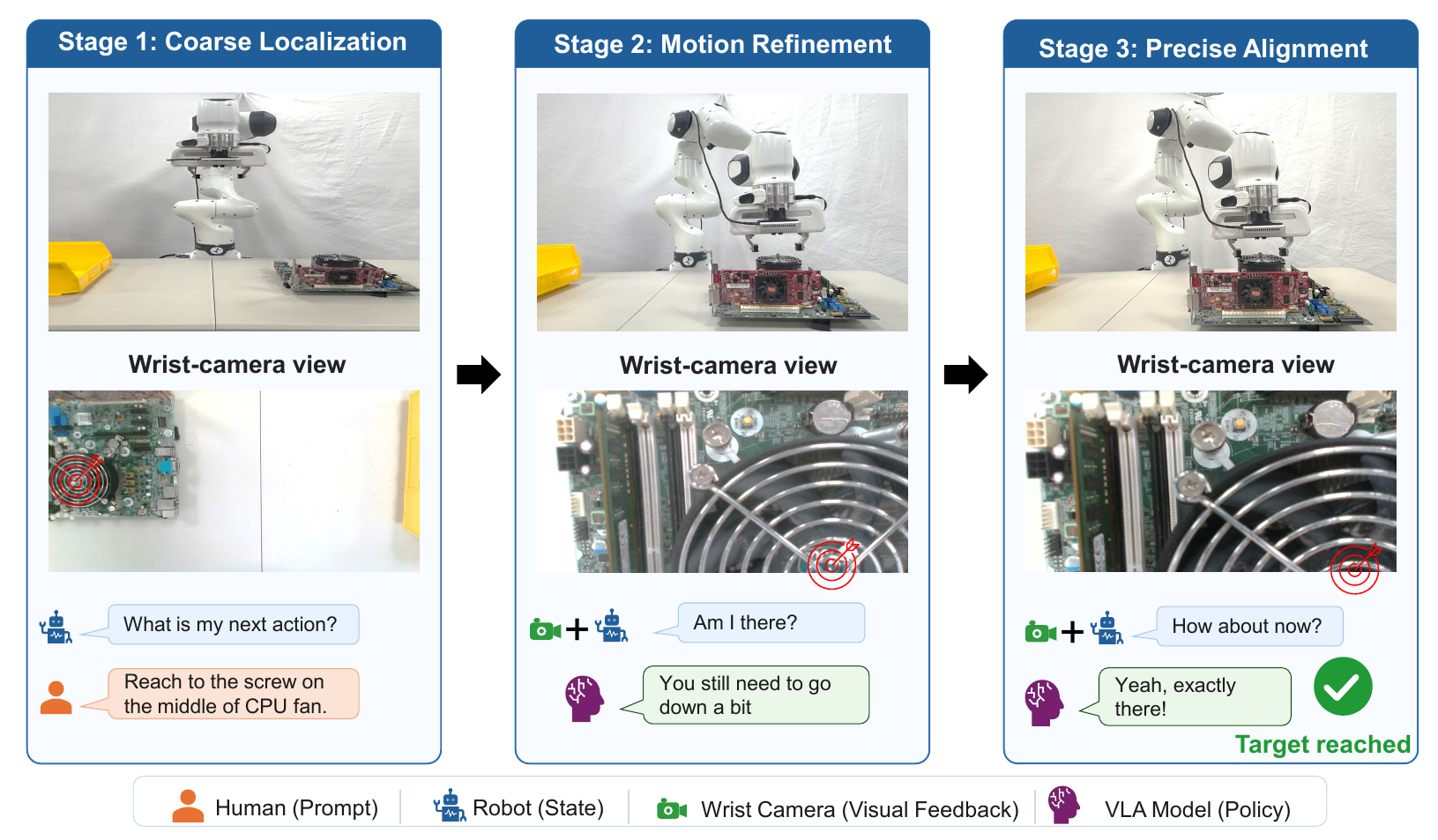}
    \caption{
    Conceptual illustration of progressive wrist-camera feedback for precise reaching.
    The robot first performs coarse target localization, then refines its motion through repeated visual feedback, and finally stops when the reach verifier confirms alignment with the target interaction region.
    }
    \label{fig:progressive_feedback}
\end{figure*}

During deployment, the policy is executed in closed loop using the live wrist-camera image and target instruction, as illustrated in Fig.~\ref{fig:progressive_feedback}. To improve safety, predicted actions are scaled and clipped before execution. For joint-space policies, the commanded joint target is computed as
\begin{equation}
\mathbf{q}^{\mathrm{cmd}}_{t+1}
=
\mathrm{clip}
\left(
\mathbf{q}_{t}
+
\mathrm{clip}(\alpha \Delta \hat{\mathbf{q}}_{t}, -\delta_{\max}, \delta_{\max}),
\mathbf{q}^{-}, \mathbf{q}^{+}
\right),
\end{equation}
where $\alpha$ is the action scale, $\delta_{\max}$ is the maximum joint update, and $\mathbf{q}^{-}, \mathbf{q}^{+}$ are the joint limits. For pose-space policies, the predicted relative end-effector motion is similarly scaled and clipped before being sent to the robot controller.

To balance speed and precision, we use a coarse-to-fine action chunking strategy. The policy predicts a short action horizon, and the number of actions executed per query is gradually reduced as the rollout progresses:
\begin{equation}
H_{k+1} \leq H_k ,
\end{equation}
where $H_k$ is the executed chunk length at rollout chunk $k$. This allows fast motion at the beginning and more frequent visual feedback near the target.

\subsection{Instruction-Conditioned Reach Verification}
Different manipulation targets impose distinct geometric requirements for
successful grasping, making a universal reaching or stopping criterion
inadequate. For example, a screw requires high precision along the vertical
approach direction, whereas thin components such as GPUs require accurate
alignment to ensure that the component lies within the gripper opening while
allowing greater tolerance along other directions. Similarly, RAM modules
permit greater positional variation along their longitudinal axis than across
their narrow width. These target-dependent tolerances motivate an
\textit{instruction-conditioned reach verifier} that complements the VLA
policy by determining whether the robot has reached a sufficiently precise
interaction configuration for the instructed target.

\begin{figure*}[t]
    \centering
    \includegraphics[width=\textwidth]{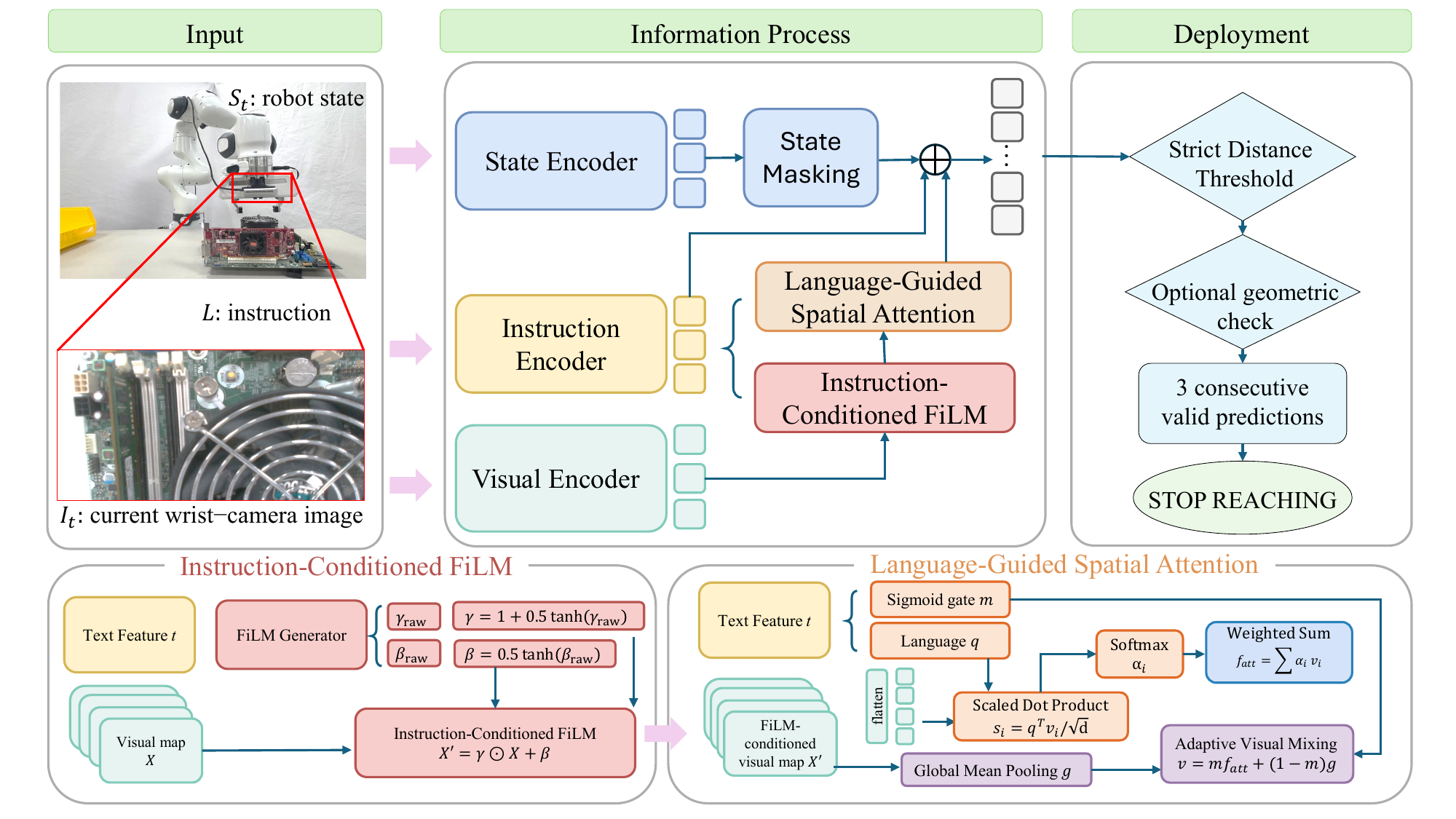}
    \caption{Overview of the instruction-conditioned reach verifier. 
    The verifier integrates the current wrist-camera observation, language instruction, 
    and robot state to estimate whether the end-effector has reached the target interaction 
    region. Language-conditioned FiLM and spatial attention adapt the visual representation 
    to the instructed target, while state masking discourages reliance on robot-state 
    shortcuts. During deployment, reaching is confirmed only after three consecutive 
    valid predictions.}
    \label{fig:reach_verifier}
\end{figure*}

We use a reach verifier, trained separately from the VLA backbone, to determine
whether the end-effector has reached the target interaction region. The overall
workflow is illustrated in Fig.~\ref{fig:reach_verifier}. Its supervision is the
Cartesian distance to the target, and training minimizes a Smooth-$L_1$
regression loss. At each policy query, the verifier estimates the target distance
from the current wrist-camera observation, target instruction, and robot state:
\begin{equation}
    \hat d_t = V(\mathbf{I}_{t}, \ell, \mathbf{s}_{t}).
\end{equation}
The three inputs are first processed by separate visual, instruction, and state
encoders. The encoded instruction conditions the visual representation through
two complementary mechanisms: instruction-conditioned FiLM applies
language-dependent modulation to the visual feature map, while language-guided
spatial attention emphasizes image regions relevant to the instructed target.
The resulting visual representation is fused with the encoded robot state for
distance prediction. During training, we apply \textit{state masking} by
randomly setting the encoded state feature to zero with a probability of 0.5.
This reduces reliance on workspace-specific cues such as absolute end-effector
height, which can become a shortcut for target proximity under a fixed workspace,
and encourages greater use of visual and instruction-conditioned cues.

At deployment, $\hat d_t < 0.65$\,cm is treated as a high-confidence reach.
To reduce premature termination caused by transient predictions, the rollout
terminates only after three consecutive high-confidence predictions, with an
optional geometric check applied before termination. This persistence rule
reduces premature stops and overshooting while allowing the robot to continue
correcting its motion until it reaches the target region. Control can then be
transferred to a downstream manipulation module.

\subsection{Downstream Grasp Execution}

Once the reach verifier confirms that the end-effector has entered the target
interaction region, control is transferred from the VLA policy to a deterministic
grasp-and-place controller. At the handoff time $t^{*}$, the current end-effector
pose is directly used as the grasp pose, without additional object-pose estimation
or grasp-pose prediction. The parallel-jaw gripper is then closed with a commanded
force of $70$\,N and a finite width tolerance to accommodate small residual
alignment errors.

After a successful grasp, the robot executes a predefined lift--transfer--place
sequence. The object is first lifted vertically from the grasp pose, transferred
horizontally to a predefined placement region, lowered, released, and the robot
then returns to its home configuration. Each Cartesian segment uses fifth-order
time scaling with quaternion interpolation for orientation and is executed under
joint impedance control.

The downstream controller is intentionally independent of the proposed reaching
framework and is kept fixed across all trials. Its purpose is to verify that the
interaction pose produced by the VLA policy and reach verifier can be directly
used for physical manipulation, without introducing an additional learned
manipulation policy.

\section{Experiments and Results}
\label{sec:result}

\subsection{Experimental Setup and Evaluation Protocol}

\begin{figure*}[t]
    \centering
    \makebox[\textwidth][c]{%
        \includegraphics[width=0.158\textwidth]{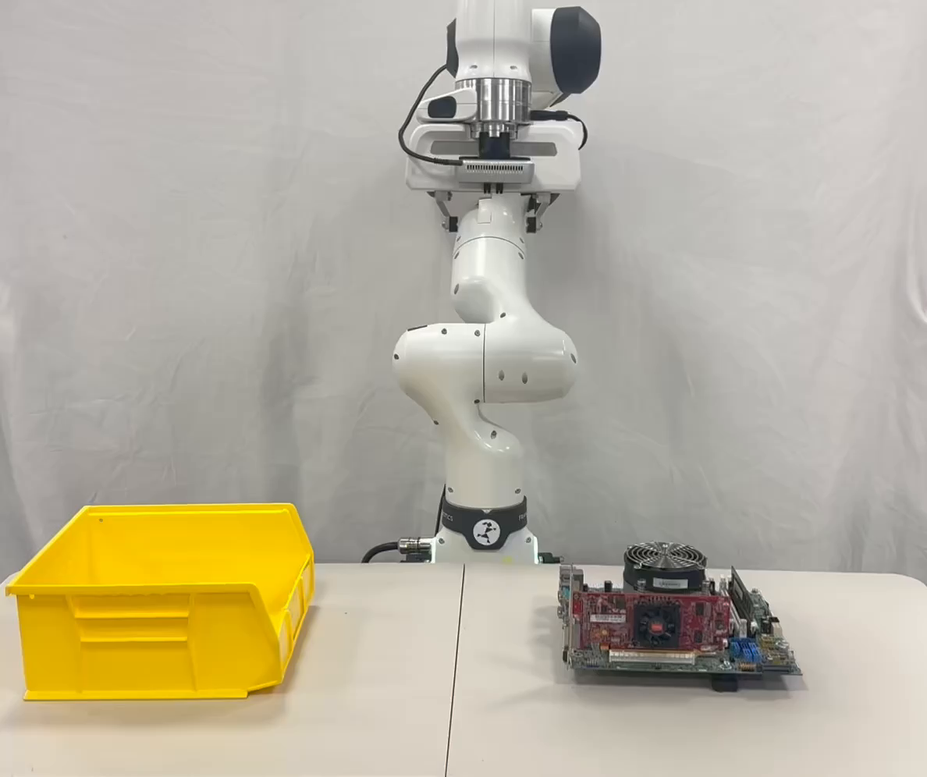}\hspace{0.002\textwidth}%
        \includegraphics[width=0.158\textwidth]{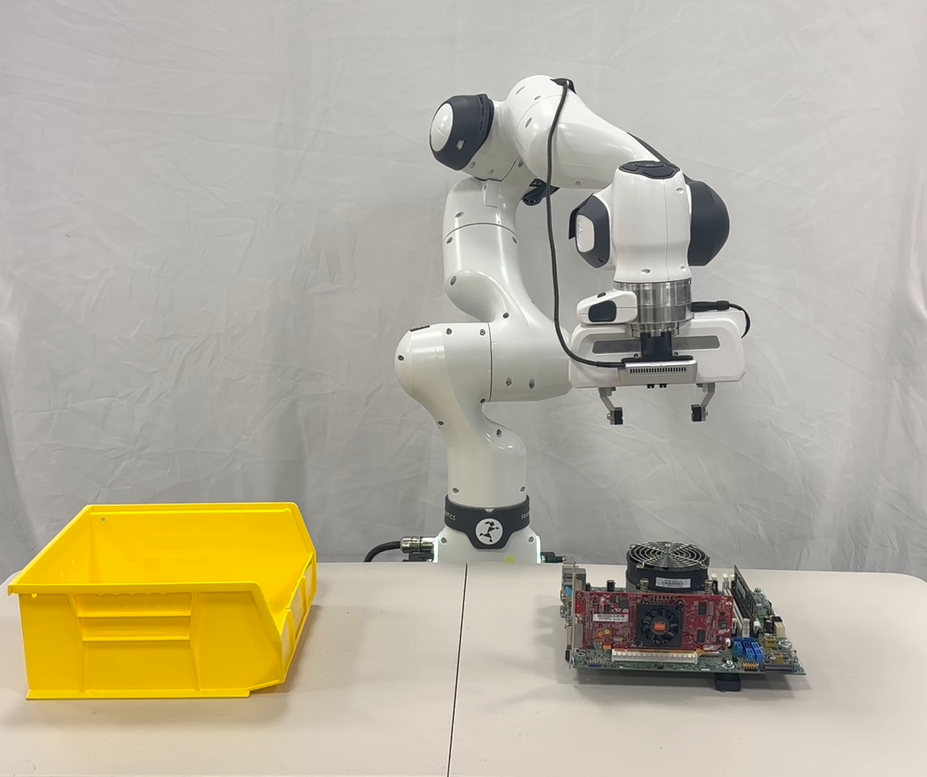}\hspace{0.002\textwidth}%
        \includegraphics[width=0.158\textwidth]{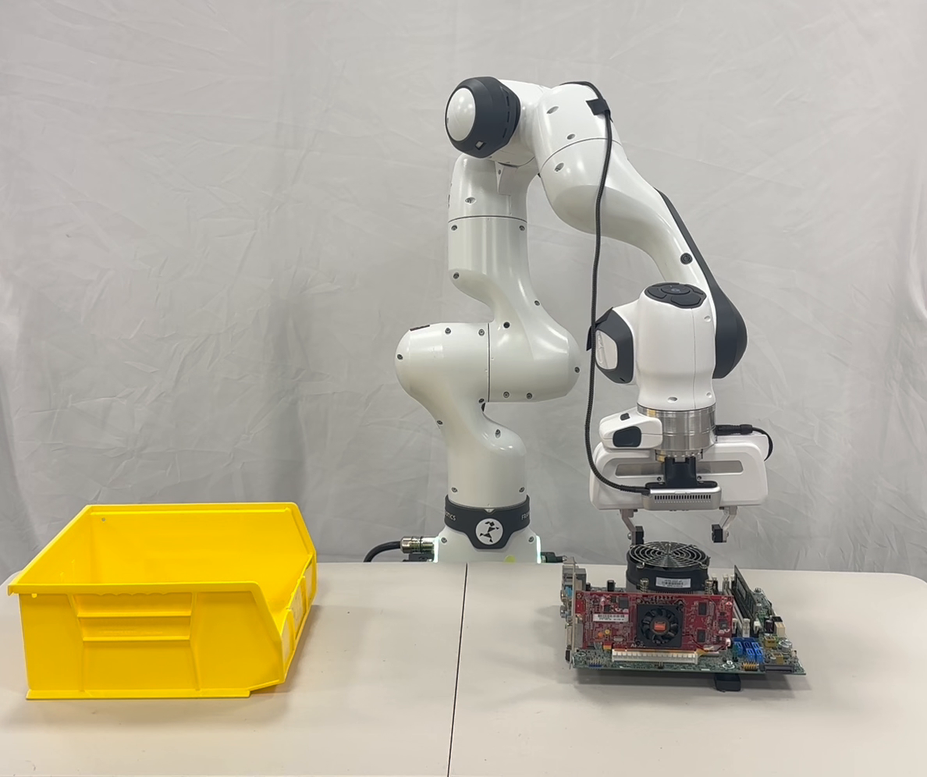}\hspace{0.002\textwidth}%
        \includegraphics[width=0.158\textwidth]{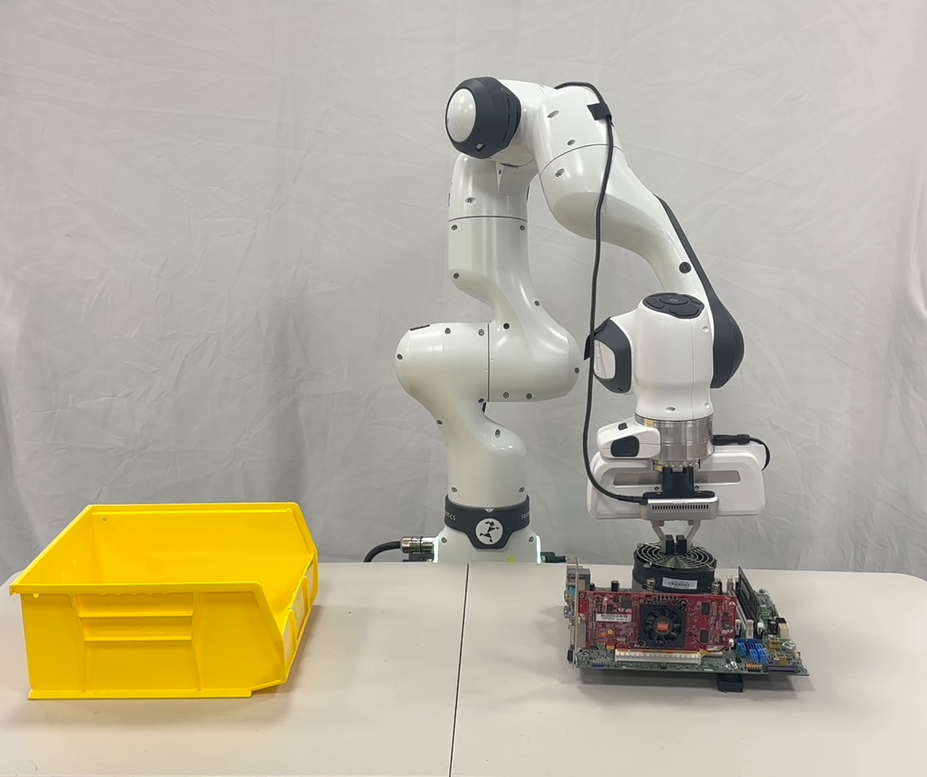}\hspace{0.002\textwidth}%
        \includegraphics[width=0.158\textwidth]{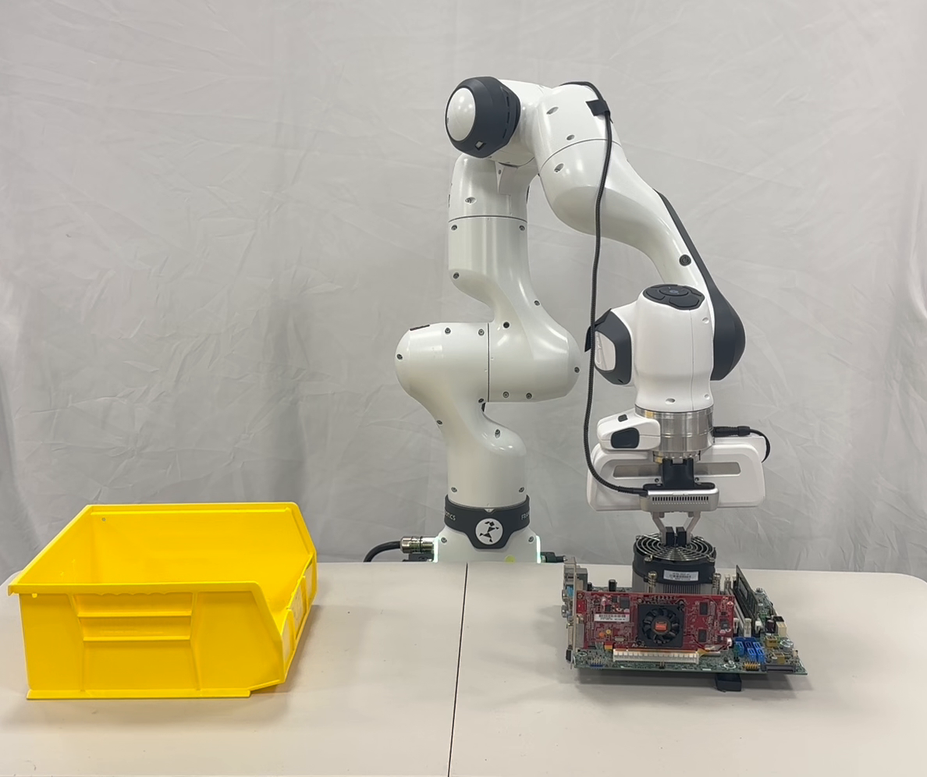}\hspace{0.002\textwidth}%
        \includegraphics[width=0.158\textwidth]{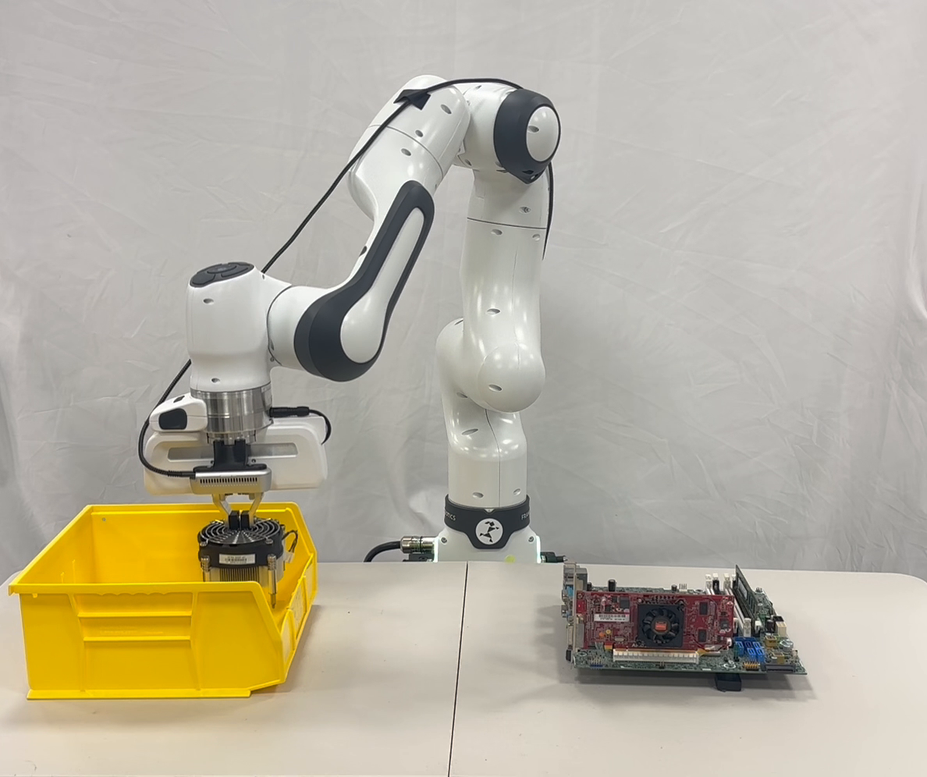}%
    }\\[2pt]
    \makebox[\textwidth][c]{%
        \includegraphics[width=0.158\textwidth]{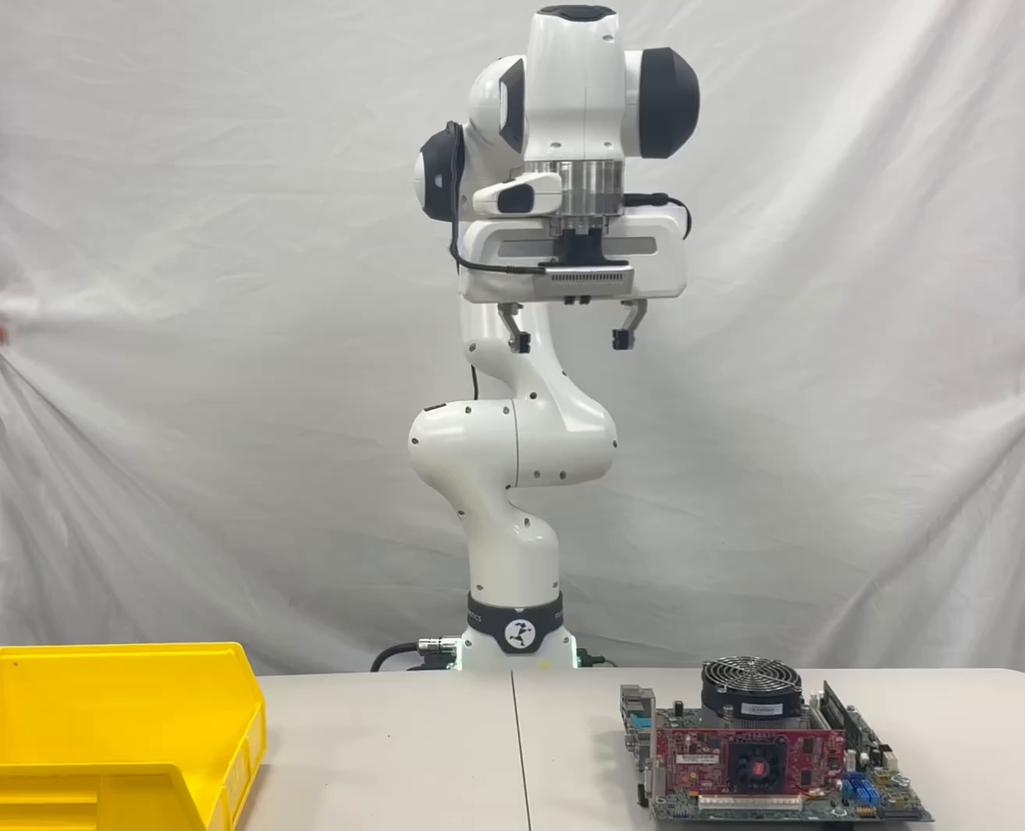}\hspace{0.002\textwidth}%
        \includegraphics[width=0.158\textwidth]{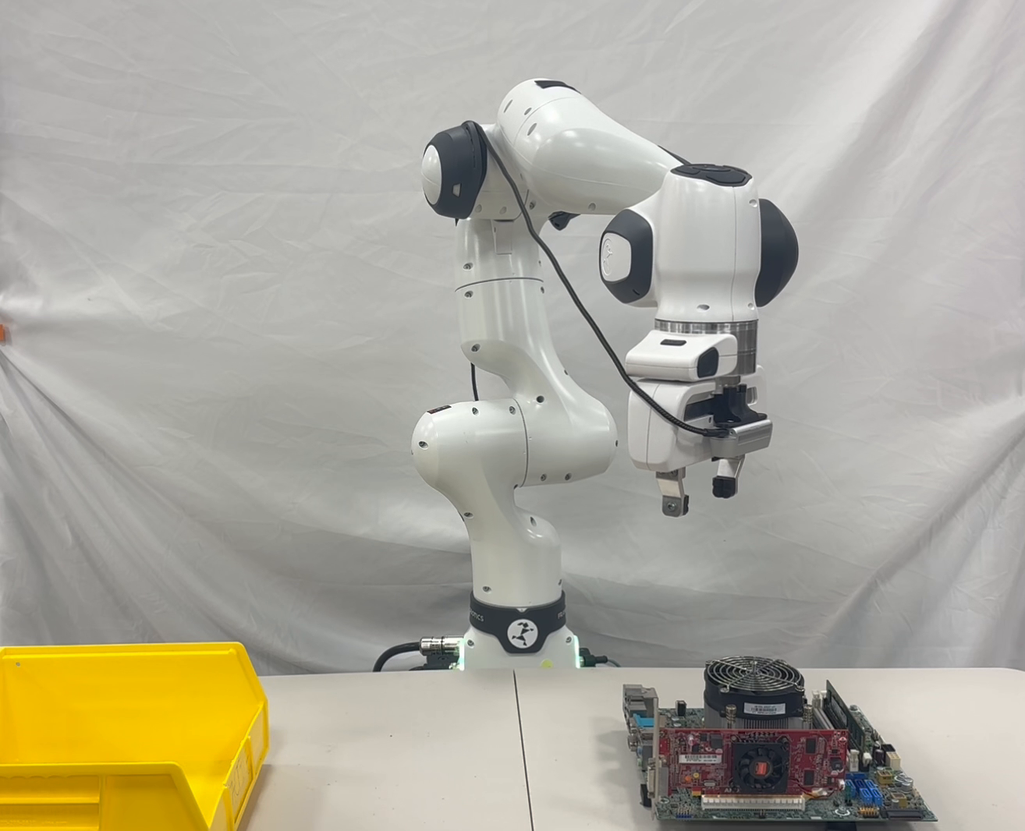}\hspace{0.002\textwidth}%
        \includegraphics[width=0.158\textwidth]{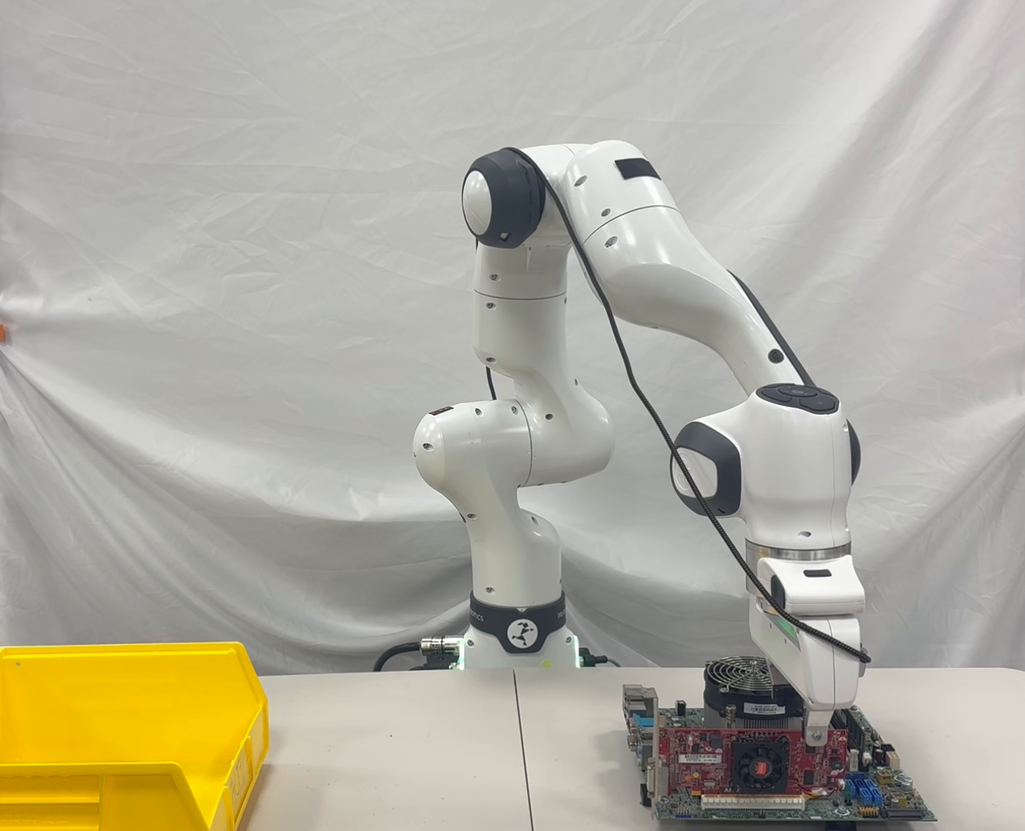}\hspace{0.002\textwidth}%
        \includegraphics[width=0.158\textwidth]{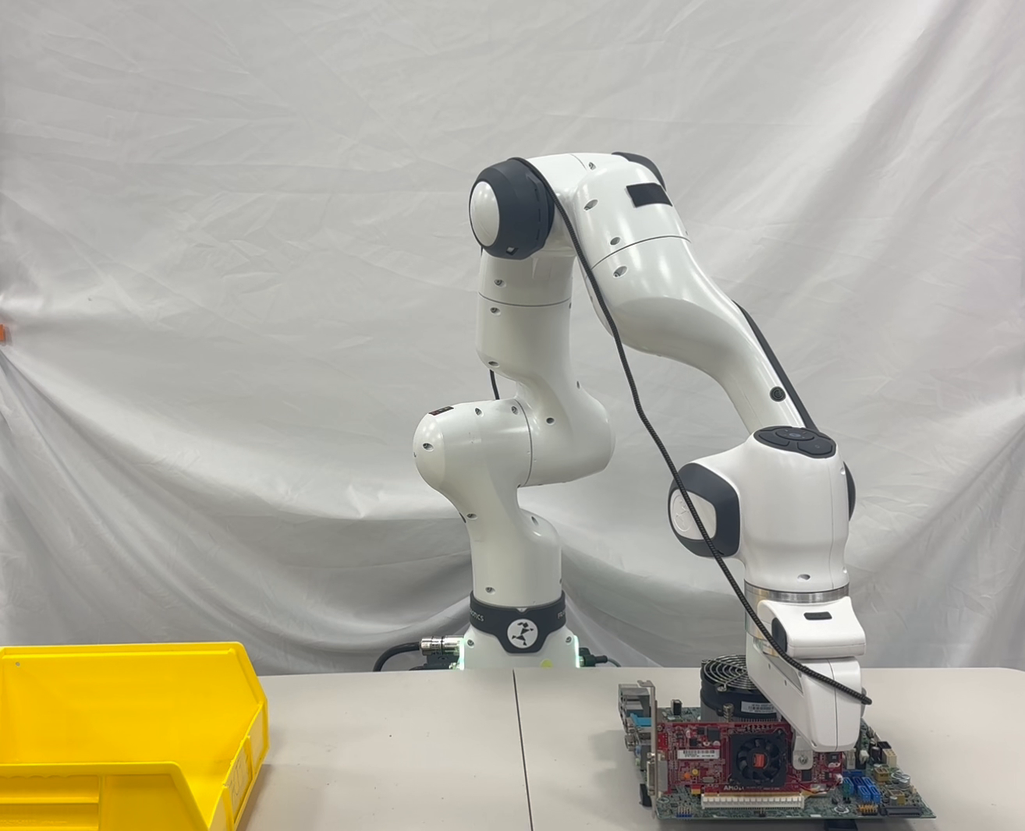}\hspace{0.002\textwidth}%
        \includegraphics[width=0.158\textwidth]{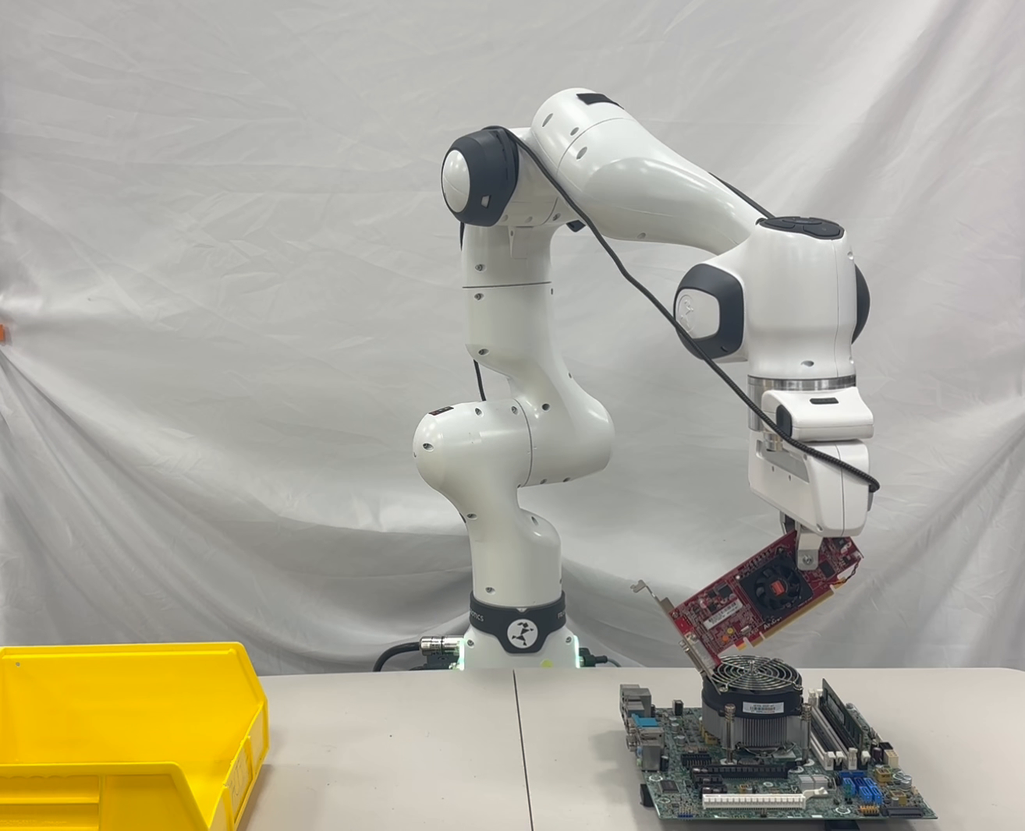}\hspace{0.002\textwidth}%
        \includegraphics[width=0.158\textwidth]{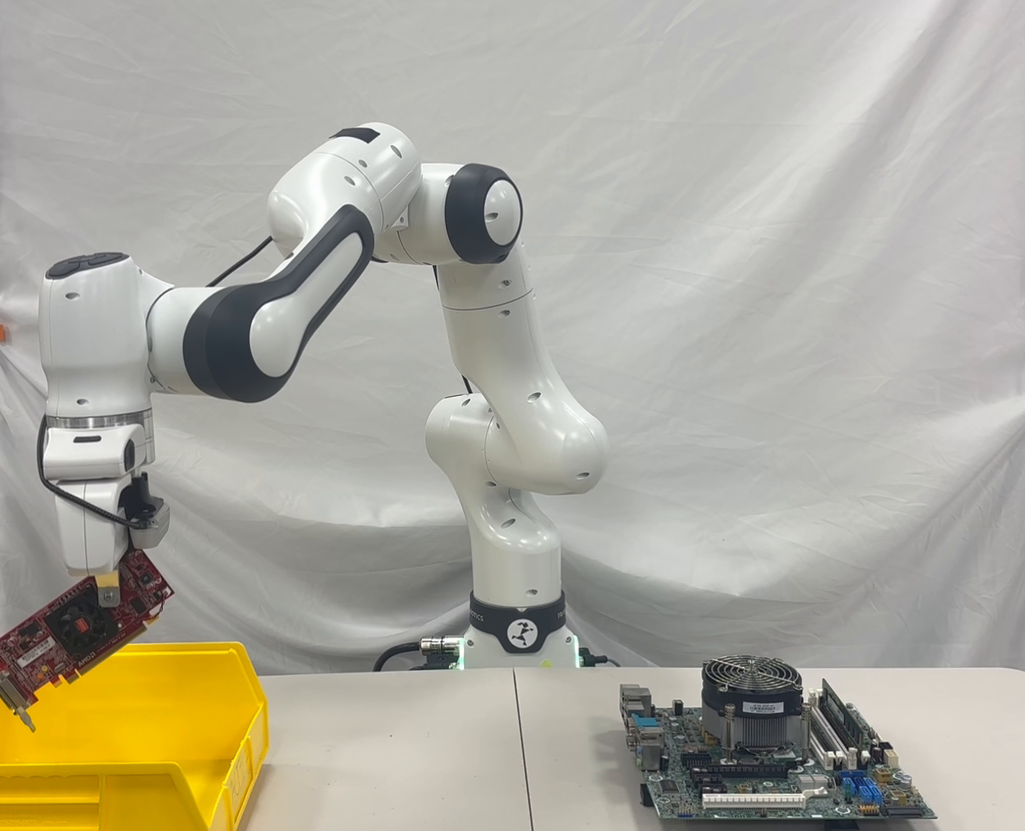}%
    }\\[2pt]
    \makebox[\textwidth][c]{%
        \includegraphics[width=0.158\textwidth]{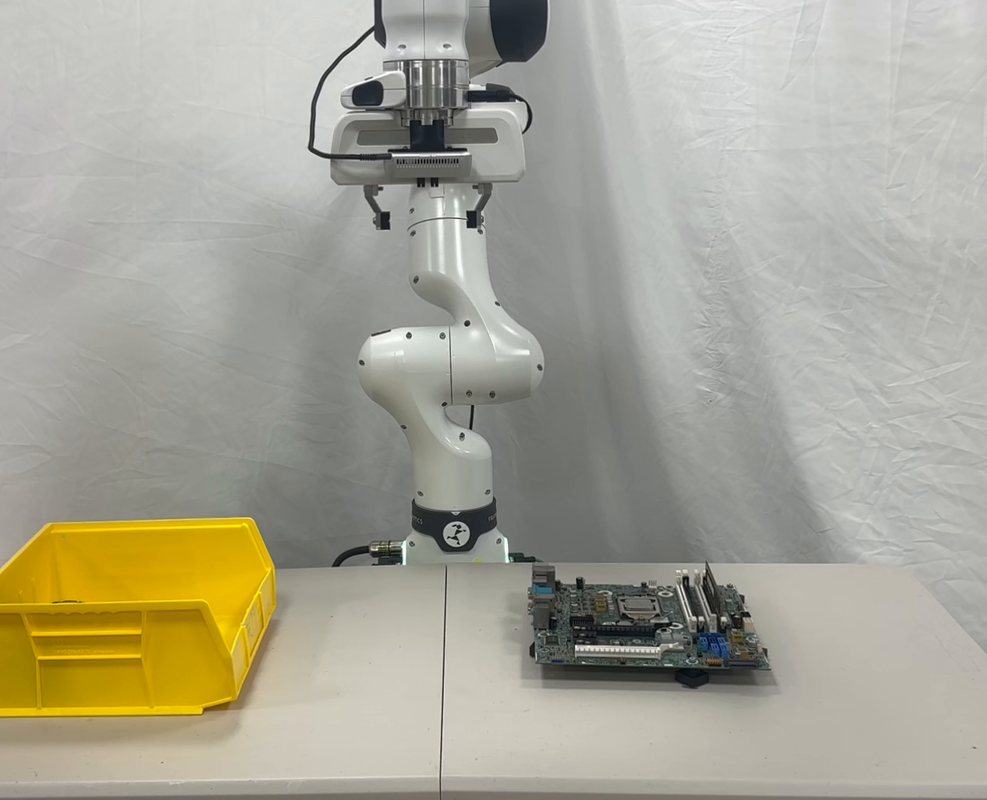}\hspace{0.002\textwidth}%
        \includegraphics[width=0.158\textwidth]{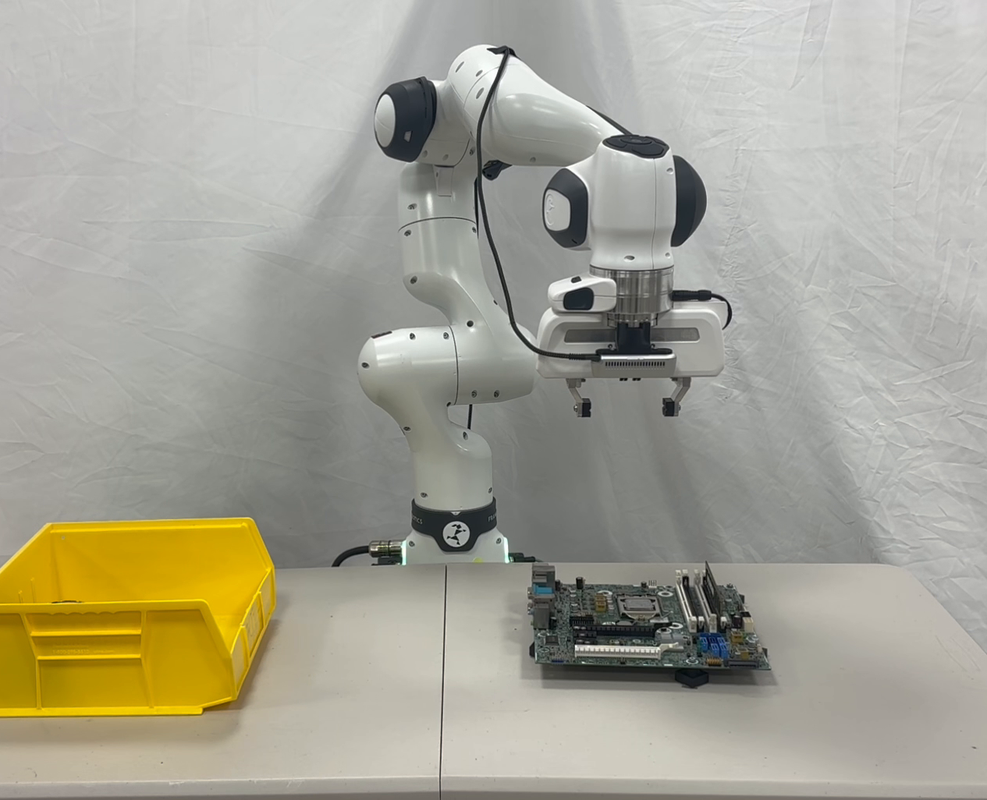}\hspace{0.002\textwidth}%
        \includegraphics[width=0.158\textwidth]{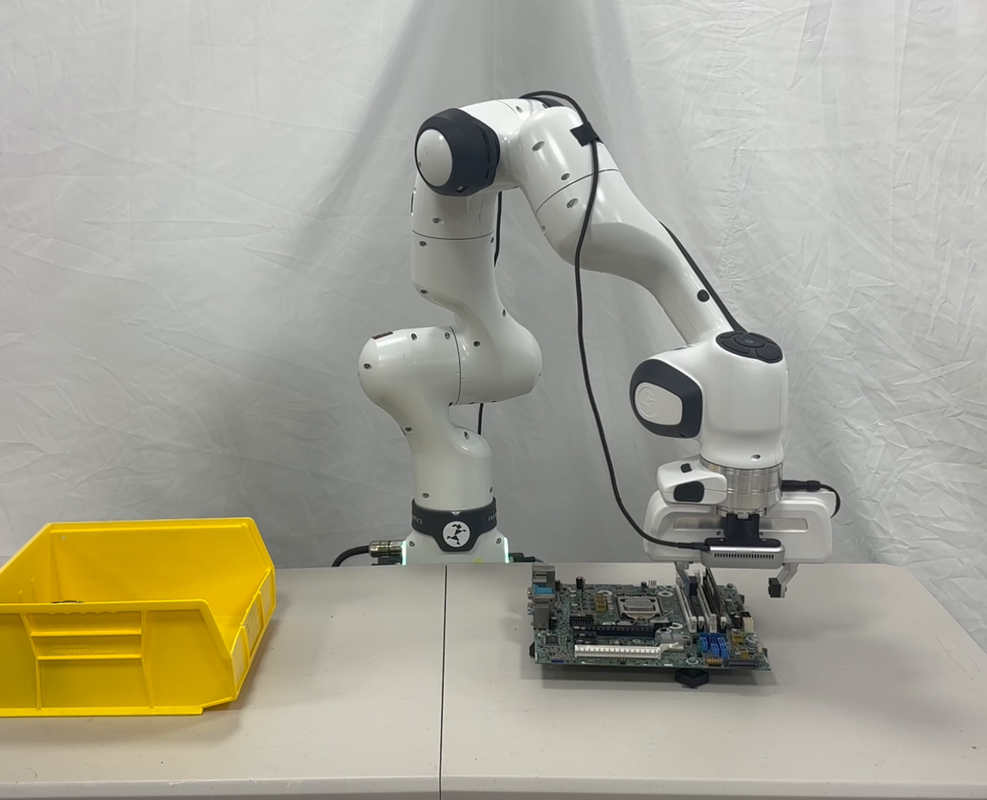}\hspace{0.002\textwidth}%
        \includegraphics[width=0.158\textwidth]{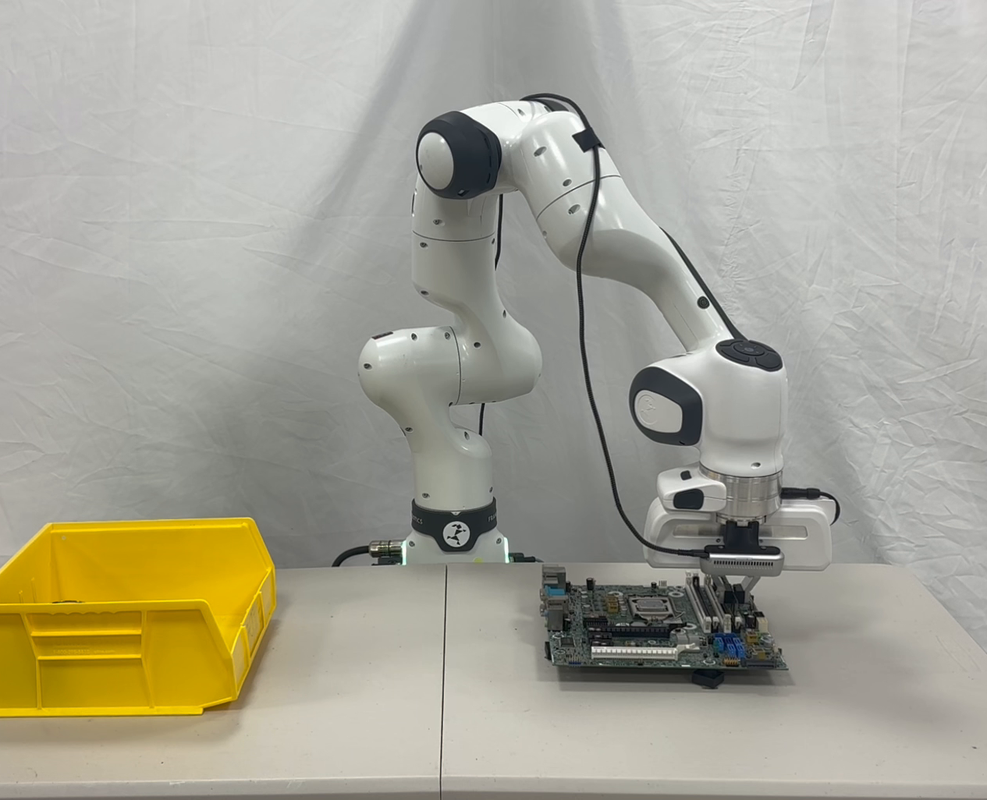}\hspace{0.002\textwidth}%
        \includegraphics[width=0.158\textwidth]{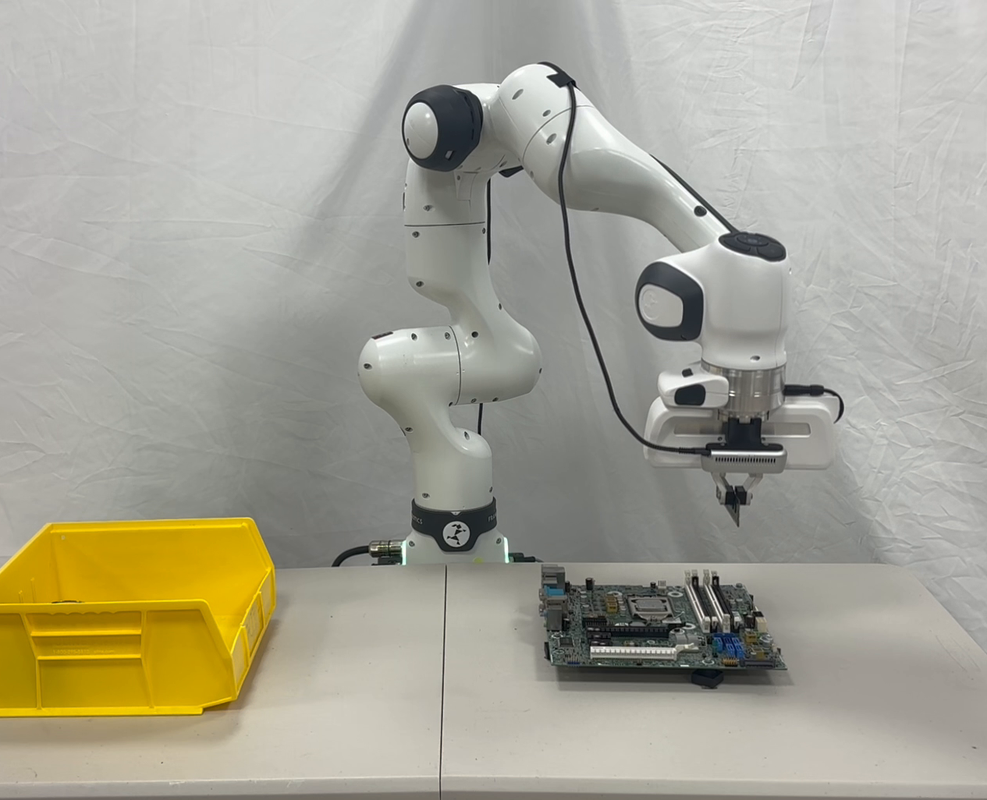}\hspace{0.002\textwidth}%
        \includegraphics[width=0.158\textwidth]{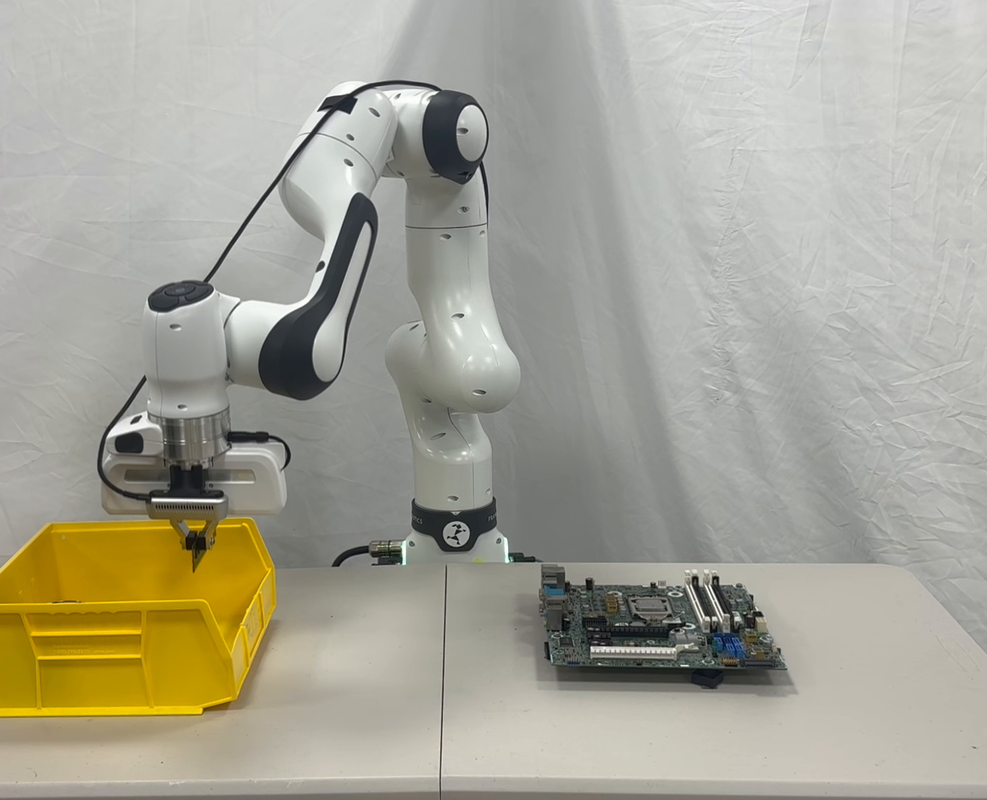}%
    }
    \caption{Qualitative progression of three real-robot reaching trajectories
    (top to bottom), each shown as six frames from the initial configuration to
    the final target interaction pose.}
    \label{fig:qualitative_trajectory}
\end{figure*}

We evaluate PrecisionReach-GS on a Franka FR3 robot with a wrist-mounted
RealSense D435i camera in a motherboard disassembly workspace.  The dataset
contains three language-conditioned targets---Screw, GPU, and RAM.  The task is
precise pre-contact reaching, where the robot must reach a specified
component-level interaction pose before downstream manipulation.  For the
teleoperation baseline, an operator controls the FR3 end effector with a
SpaceMouse while observing synchronized wrist- and side-camera streams.  That
baseline is trained and evaluated with both views, providing a stronger
sensor-rich setting.  In contrast, PrecisionReach-GS is trained and evaluated
using only wrist-camera observations, with 300, 900, and 1,500 generated
trajectories.  Experiments use both OpenVLA-OFT and $\pi_{0.5}$ under the same
real-robot protocol.
A set of qualitative reaching trajectories is shown in
Fig.~\ref{fig:qualitative_trajectory}.

We report two real-robot success metrics that isolate the learned reaching
stage. Coarse reach counts arrival in the target neighborhood. Precise reach
requires arrival at the target interaction region; we validate that terminal
pose by executing the same downstream grasp-and-lift routine in every trial.
Thus, the metric measures whether reaching supplies a manipulation-ready pose
rather than asking the VLA to learn contact. The real-robot tests are conducted
after physically repositioning the motherboard relative to the scene used for
3DGS reconstruction and generation. The resulting target-location changes test
moderate scene variation instead of fixed-pose replay. The task geometry also
makes error tolerance anisotropic: GPU and RAM targets permit somewhat more
lateral error along their long axes, with additional margin from the gripper
width, whereas approach-direction error remains tightly constrained for all
targets.

\subsection{Main Real-Robot Performance and Data Efficiency}

\begin{figure*}[t]
    \centering
    \includegraphics[width=\textwidth]{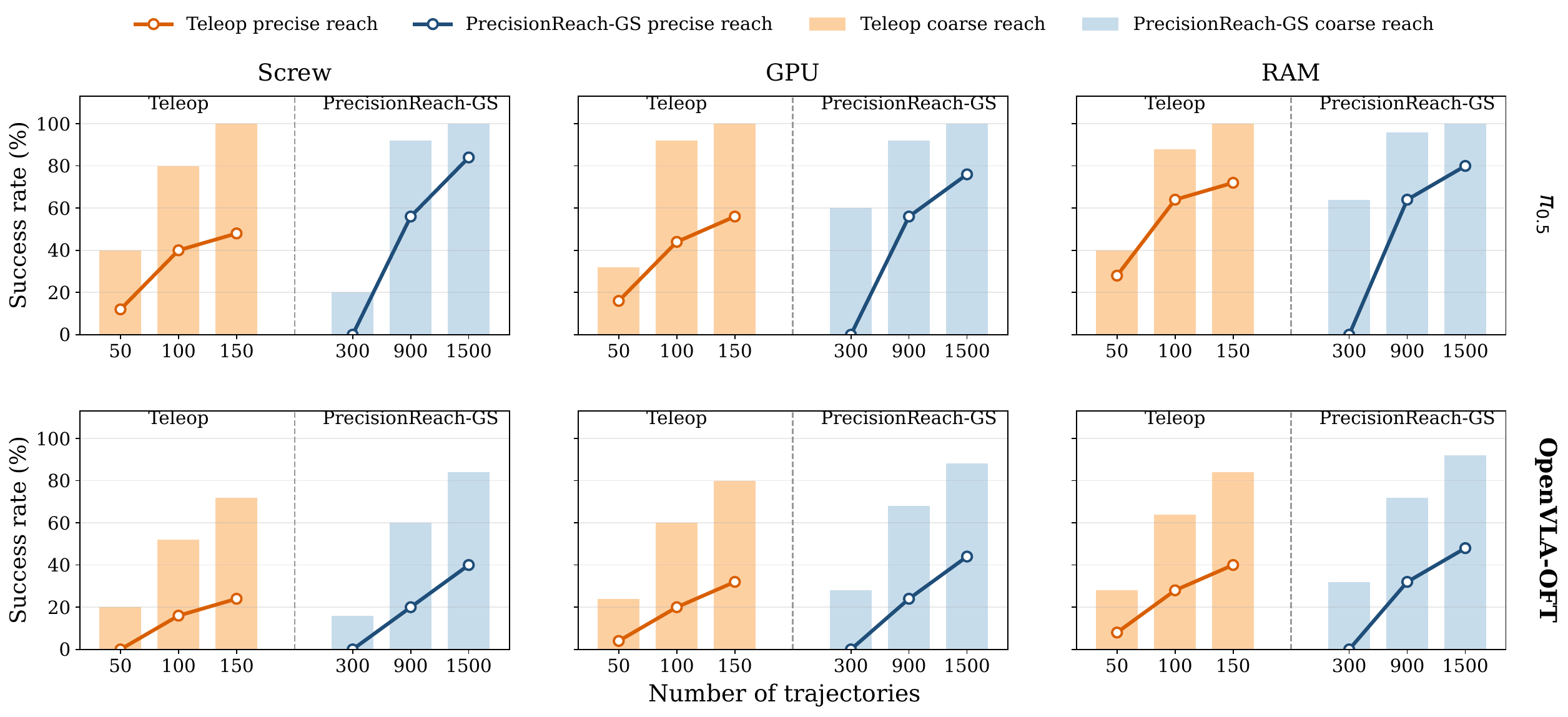}
    \caption{
    Real-robot reaching performance on a physically repositioned motherboard workspace that differs from the scene used for 3DGS reconstruction and data generation. Detailed trial counts are reported in Table~\ref{tab:success_rate_scale}.
    }
    \label{fig:success_rate_pi05_openvla}
\end{figure*}

As shown in
Fig.~\ref{fig:success_rate_pi05_openvla}, the improvement is especially clear
for $\pi_{0.5}$, where 900 and 1,500 PrecisionReach-GS trajectories achieve
strong precise-reaching performance across all three targets.  OpenVLA-OFT
shows the same increasing trend but lower overall success.  Notably,
PrecisionReach-GS uses only the wrist camera, whereas the original teleoperation
baseline uses both wrist and side views. Table~\ref{tab:success_rate_scale}
provides the corresponding per-target trial counts and makes the data-scale
trend explicit.

At the largest generated-data scale, $\pi_{0.5}$ achieves 21/25, 19/25,
and 20/25 precise successes for screw, GPU, and RAM, respectively
(Table~\ref{tab:success_rate_scale}). The GPU target is slightly more
challenging because it is larger and taller than the other targets and requires
greater wrist rotation; near the target, that rotation can occasionally cause a
collision. Across both policy backbones, however, increasing the number of
generated trajectories consistently improves coarse and precise reaching.

\begin{table*}[!t]
\centering
\caption{
Real-robot reaching performance under different amounts of teleoperation and PrecisionReach-GS training data.
Each entry reports successful trials out of 25.
Coarse reach measures arrival in the target neighborhood; precise reach additionally requires a terminal pose from which the fixed grasp-and-lift routine succeeds.
}
\label{tab:success_rate_scale}
\small
\setlength{\tabcolsep}{3pt}
\renewcommand{\arraystretch}{1.12}
\begin{tabular}{@{}lll*{6}{C{0.075\textwidth}}@{}}
\toprule
\textbf{Policy}
& \textbf{Target}
& \textbf{Metric}
& \multicolumn{3}{c}{\textbf{Teleop Trajectories}}
& \multicolumn{3}{c}{\textbf{PrecisionReach-GS Trajectories}} \\
\cmidrule(lr){4-6}
\cmidrule(lr){7-9}
& &
& \textbf{50}
& \textbf{100}
& \textbf{150}
& \textbf{300}
& \textbf{900}
& \textbf{1500} \\
\midrule

\multirow{6}{*}{$\pi_{0.5}$}
& \multirow{2}{*}{Screw}
& Coarse reach
& 10/25 & 20/25 & 25/25
& 5/25 & 23/25 & 25/25 \\
&
& Precise reach
& 3/25 & 10/25 & 12/25
& 0/25 & 14/25 & 21/25 \\

& \multirow{2}{*}{GPU}
& Coarse reach
& 8/25 & 23/25 & 25/25
& 15/25 & 23/25 & 25/25 \\
&
& Precise reach
& 4/25 & 11/25 & 14/25
& 0/25 & 14/25 & 19/25 \\

& \multirow{2}{*}{RAM}
& Coarse reach
& 10/25 & 22/25 & 25/25
& 16/25 & 24/25 & 25/25 \\
&
& Precise reach
& 7/25 & 16/25 & 18/25
& 0/25 & 16/25 & 20/25 \\

\midrule

\multirow{6}{*}{OpenVLA-OFT}
& \multirow{2}{*}{Screw}
& Coarse reach
& 5/25 & 13/25 & 18/25
& 4/25 & 15/25 & 21/25 \\
&
& Precise reach
& 0/25 & 4/25 & 6/25
& 0/25 & 5/25 & 10/25 \\

& \multirow{2}{*}{GPU}
& Coarse reach
& 6/25 & 15/25 & 20/25
& 7/25 & 17/25 & 22/25 \\
&
& Precise reach
& 1/25 & 5/25 & 8/25
& 0/25 & 6/25 & 11/25 \\

& \multirow{2}{*}{RAM}
& Coarse reach
& 7/25 & 16/25 & 21/25
& 8/25 & 18/25 & 23/25 \\
&
& Precise reach
& 2/25 & 7/25 & 10/25
& 0/25 & 8/25 & 12/25 \\

\bottomrule
\end{tabular}
\end{table*}

Table~\ref{tab:data_efficiency} compares the data-acquisition cost and trajectory characteristics of teleoperation and PrecisionReach-GS. 
Collecting 50--150 teleoperation demonstrations requires 90--200 minutes of continuous human control, whereas PrecisionReach-GS only requires less than 10 minutes of active human capture to generate up to 1,500 trajectories after reconstruction. 
The reported generation time is measured for a single generation window; because each process uses less than 1 GB of GPU memory, multiple windows can be run in parallel depending on available GPU resources. 
This makes the proposed pipeline lightweight compared with simulator-based data generation pipelines that require high-memory GPU simulation environments. 
PrecisionReach-GS also provides smoother supervision, as reflected by its much lower RMS jerk.
Table~\ref{tab:trajectory_smoothness} expands this comparison beyond the mean:
the generated trajectories reduce median, 95th-percentile, and maximum jerk as
well, showing that the improvement is not driven by a small number of episodes.
\begin{table*}[t]
\centering
\caption{
Data acquisition cost and trajectory smoothness for teleoperation and PrecisionReach-GS.
For PrecisionReach-GS, total time is measured for a single generation window.
RMS jerk is reported in $\mathrm{m/s^3}$; lower is better.
Detailed jerk statistics are reported in Table~\ref{tab:trajectory_smoothness}.
}

\label{tab:data_efficiency}
\setlength{\tabcolsep}{3.5pt}
\renewcommand{\arraystretch}{1.12}
\begin{tabular}{lccccc}
\hline
\textbf{Method} 
& \textbf{\# Traj.} 
& \textbf{Real Capture} 
& \makecell{\textbf{Active Human Time}} 
& \textbf{Total Time} \\
\hline
\multirow{3}{*}{Teleoperation}
& 50  
& 50 demos  
& $\sim$90 min 
& $\sim$90 min \\

& 100 
& 100 demos 
& $\sim$150 min 
& $\sim$150 min 
& \\

& 150 
& 150 demos 
& $\sim$200 min 
& $\sim$200 min  
& \\
\hline
\multirow{3}{*}{PrecisionReach-GS}
& 300  
& 4 videos  
& $<5$ min 
& $\sim$45 min \\

& 900  
& 12 videos 
& $<10$ min 
& $\sim$120 min 
& \\

& 1500 
& 12 videos 
& $<10$ min 
& $\sim$200 min 
& \\
\hline
\end{tabular}
\end{table*}

\begin{table*}[!t]
    \centering
    \caption{
    Initial-view robustness under different generated data scales and target components.
    Starting views are grouped by the amount of workspace context visible in the wrist-camera observation.
    }
    \label{tab:initial_view_robustness}
    \setlength{\tabcolsep}{4pt}
    \renewcommand{\arraystretch}{1.12}
    \begin{tabular}{llcc|cc}
    \hline
    \textbf{Target}
    & \textbf{Initial Visual Context}
    & \multicolumn{2}{c|}{\textbf{300 Traj.}}
    & \multicolumn{2}{c}{\textbf{1500 Traj.}} \\
    \cline{3-6}
    &
    & \textbf{Coarse SR} & \textbf{Precise SR}
    & \textbf{Coarse SR} & \textbf{Precise SR} \\
    \hline
    \multirow{3}{*}{Screw}
    & Full workspace context     & 4/10 & 0/10 & 10/10 & 10/10 \\
    & Limited workspace context  & 1/10 & 0/10 & 9/10  & 9/10  \\
    & No workspace context       & 0/10 & 0/10 & 6/10  & 2/10  \\
    \hline
    \multirow{3}{*}{GPU}
    & Full workspace context     & 8/10 & 0/10 & 10/10 & 9/10 \\
    & Limited workspace context  & 6/10 & 0/10 & 9/10  & 9/10 \\
    & No workspace context       & 2/10 & 0/10 & 5/10  & 3/10 \\
    \hline
    \multirow{3}{*}{RAM}
    & Full workspace context     & 8/10 & 0/10 & 10/10 & 10/10 \\
    & Limited workspace context  & 6/10 & 0/10 & 10/10 & 9/10  \\
    & No workspace context       & 3/10 & 0/10 & 6/10  & 3/10  \\
    \hline
    \textbf{Average}
    & Full workspace context     & 20/30 & 0/30 & 30/30 & 29/30 \\
    \textbf{Average}
    & Limited workspace context  & 13/30 & 0/30 & 28/30 & 27/30 \\
    \textbf{Average}
    & No workspace context       & 5/30  & 0/30 & 17/30 & 8/30  \\
    \hline
    \end{tabular}
    \end{table*}    

\begin{table}[!t]
\centering
\caption{
Trajectory smoothness comparison between teleoperation and PrecisionReach-GS.
RMS jerk is reported in $\mathrm{m/s^3}$; lower values indicate smoother trajectories.
}
\label{tab:trajectory_smoothness}
\scriptsize
\setlength{\tabcolsep}{4.5pt}
\renewcommand{\arraystretch}{1.1}

\begin{tabular}{lccccc}
\hline
\textbf{Method} & \textbf{Mean} $\downarrow$
& \textbf{Median} $\downarrow$
& \textbf{P95} $\downarrow$
& \textbf{Max} $\downarrow$ \\
\hline
Teleoperation      & 3.2657 & 3.2532 & 4.1228 & 4.5666 \\
PrecisionReach-GS  & 0.2353 & 0.1683 & 0.6969 & 0.9853 \\
\hline
\end{tabular}

\renewcommand{\arraystretch}{1.0}
\end{table}

\subsection{Robustness to the Initial Wrist-Camera View}
\label{sec:initial_view_robustness}

We next test whether data scale improves recovery when the initial wrist image
offers weak task context. Starting views are grouped into full workspace
context, limited workspace context, and no workspace context. In the most
challenging no-context setting, the motherboard is completely outside the
initial wrist-camera field of view, although peripheral environmental cues,
such as the table edge, may remain visible.

\begin{figure*}[t]
    \centering
    \setlength{\tabcolsep}{2pt}
    \begin{tabular}{c c c c c c}
        \adjustbox{valign=m}{\rotatebox[origin=c]{90}{\textbf{Robot pose}}} &
        \adjustbox{valign=m}{\includegraphics[width=0.18\textwidth]{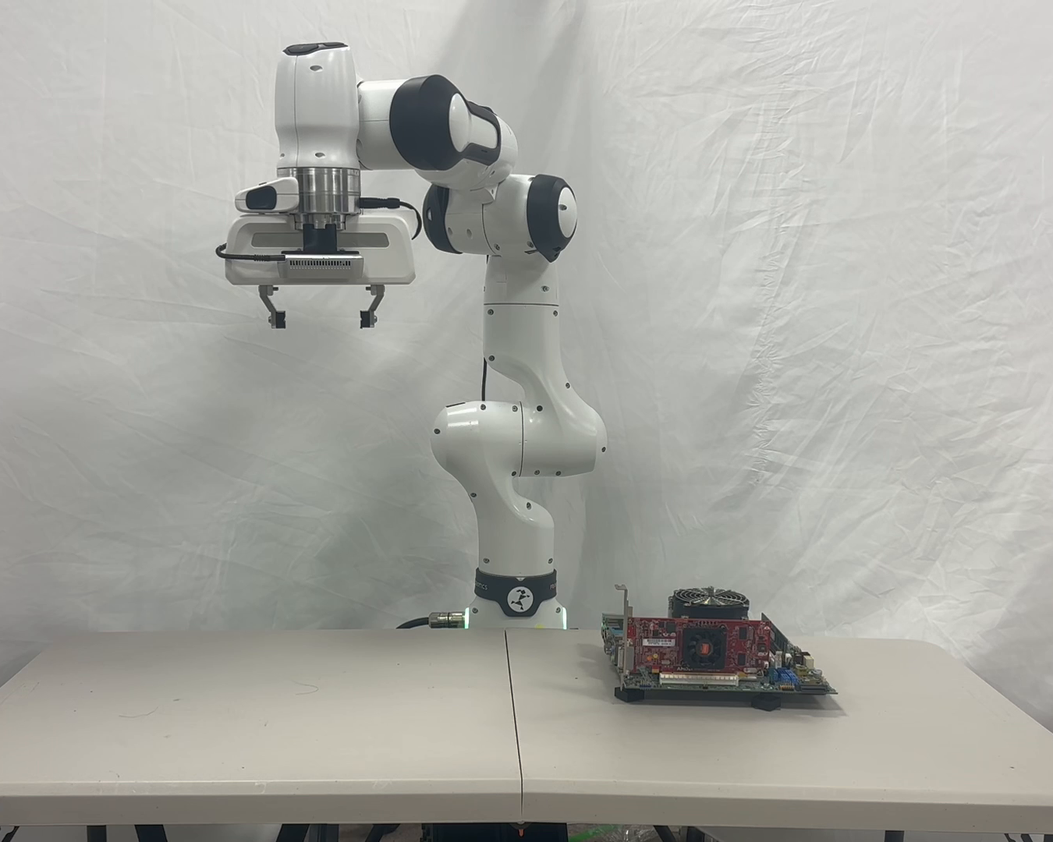}} &
        \adjustbox{valign=m}{\includegraphics[width=0.18\textwidth]{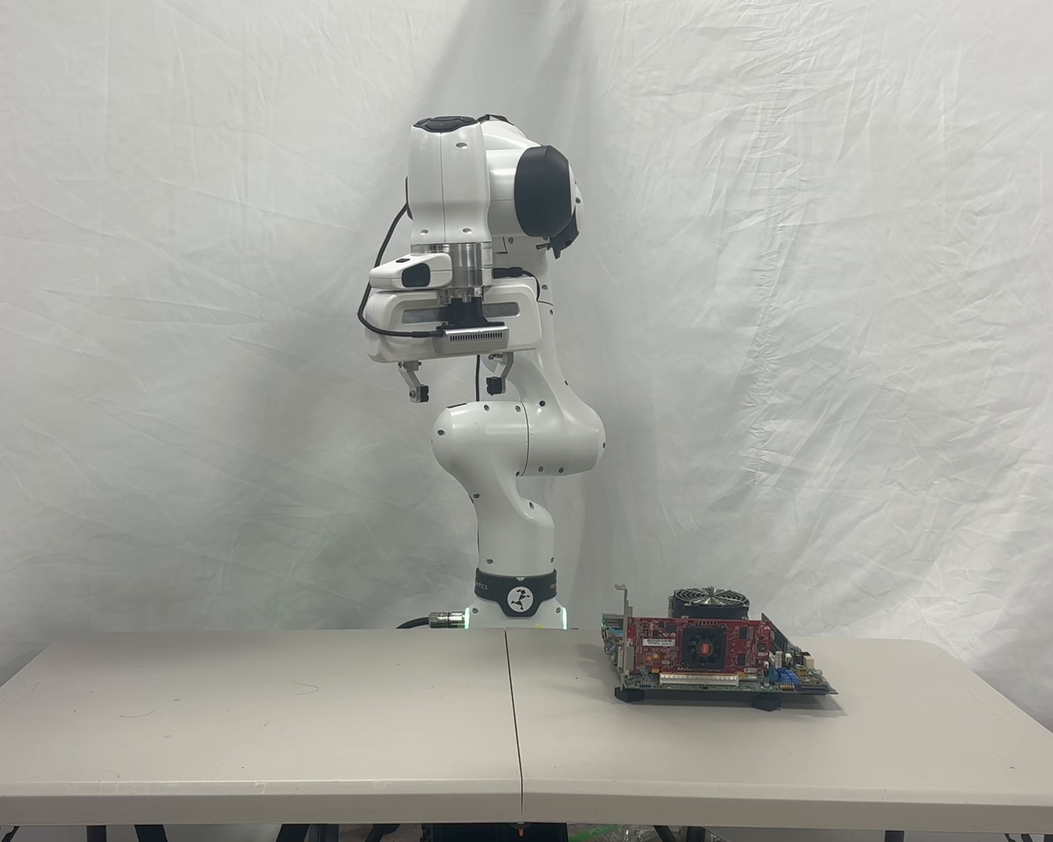}} &
        \adjustbox{valign=m}{\includegraphics[width=0.18\textwidth]{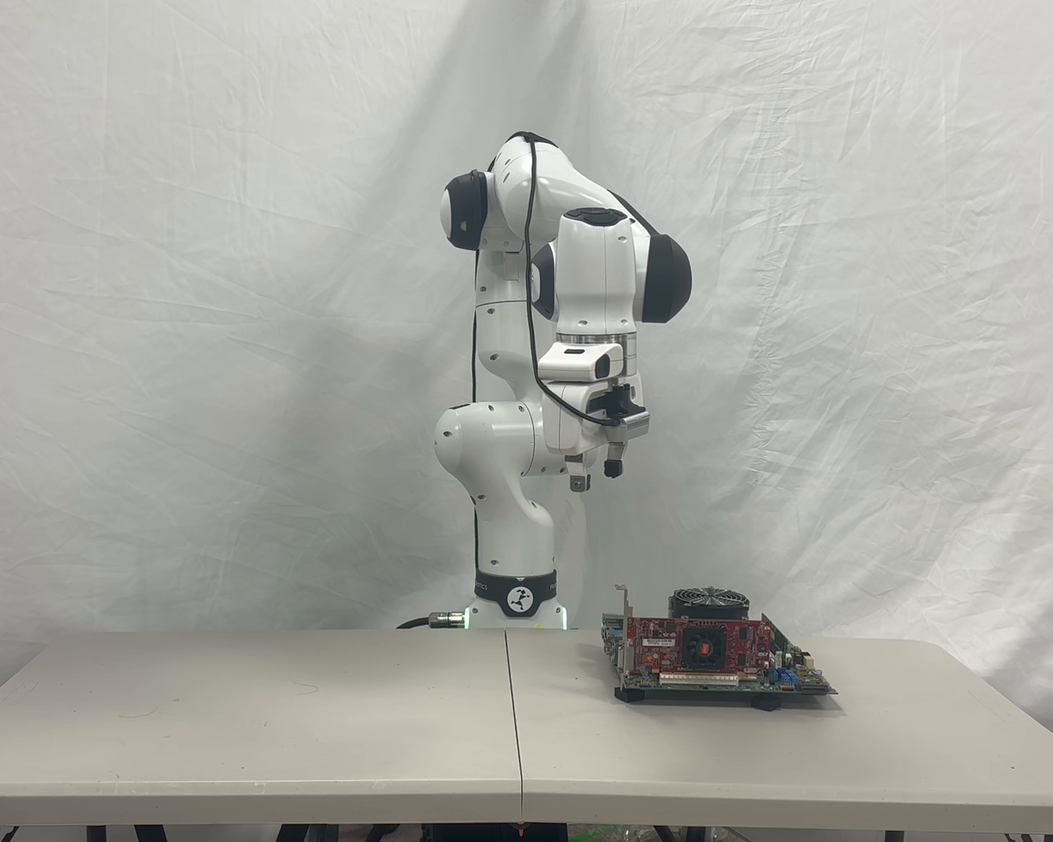}} &
        \adjustbox{valign=m}{\includegraphics[width=0.18\textwidth]{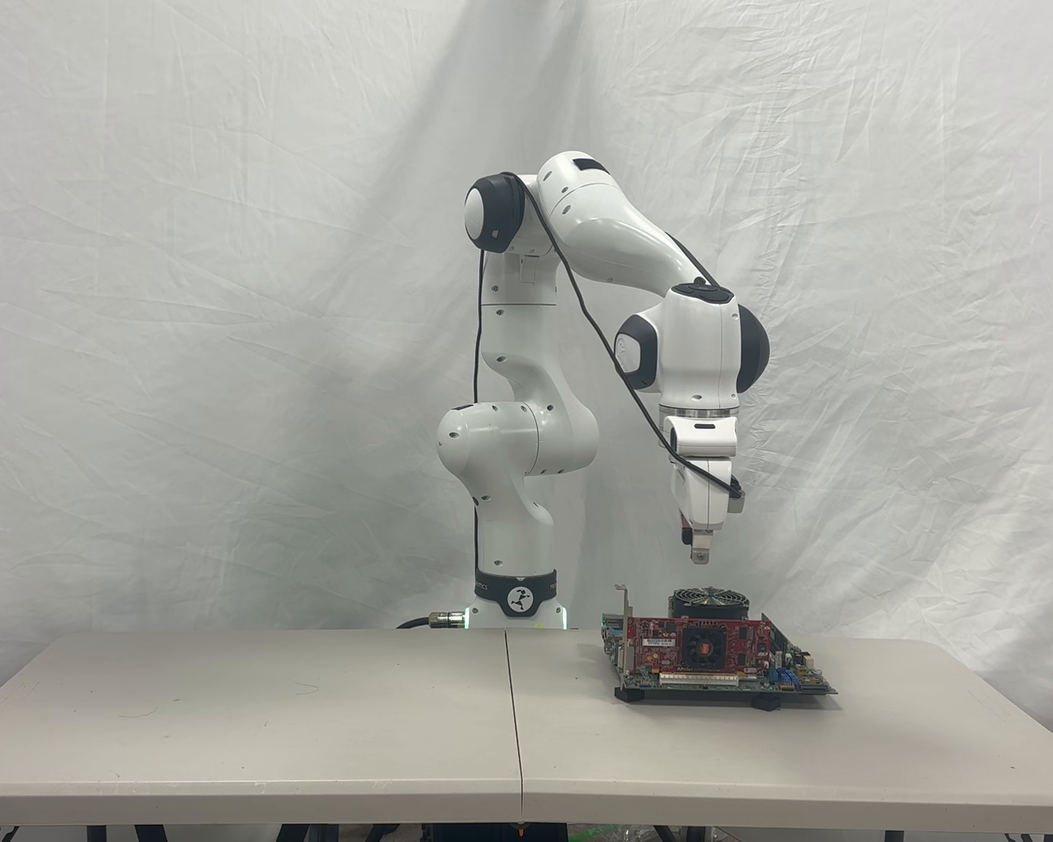}} &
        \adjustbox{valign=m}{\includegraphics[width=0.18\textwidth]{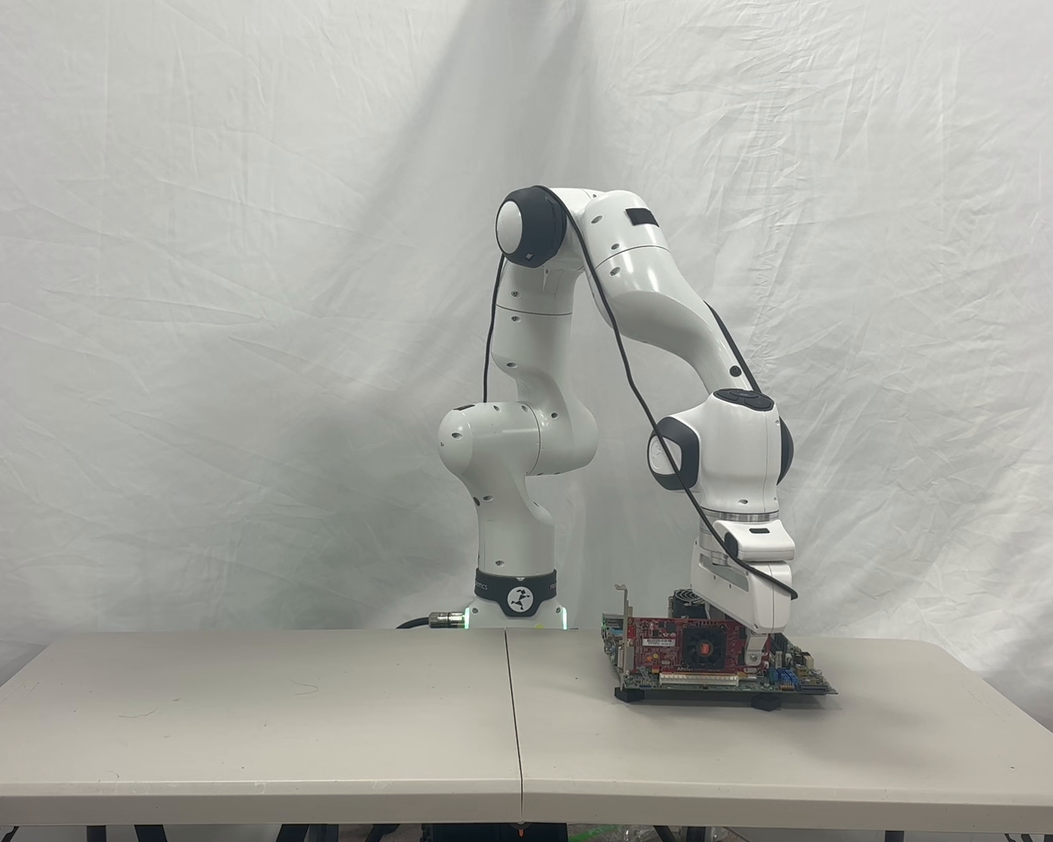}}
        \\
        \noalign{\vskip 3pt}
        \adjustbox{valign=m}{\rotatebox[origin=c]{90}{\textbf{Wrist view}}} &
        \adjustbox{valign=m}{\includegraphics[width=0.18\textwidth]{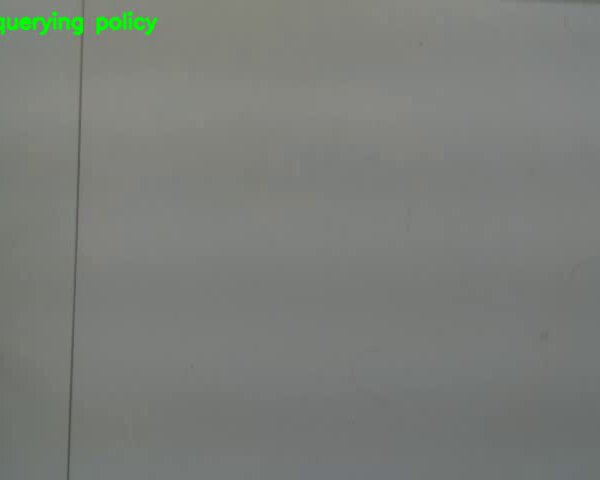}} &
        \adjustbox{valign=m}{\includegraphics[width=0.18\textwidth]{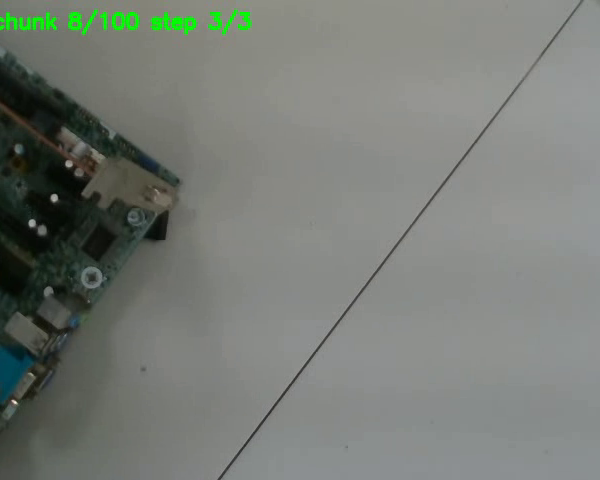}} &
        \adjustbox{valign=m}{\includegraphics[width=0.18\textwidth]{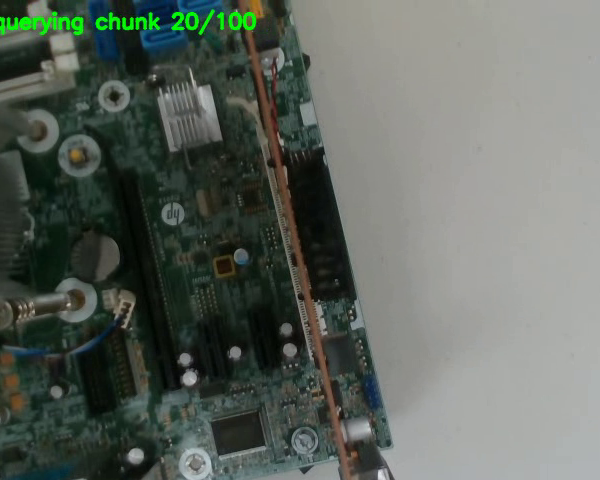}} &
        \adjustbox{valign=m}{\includegraphics[width=0.18\textwidth]{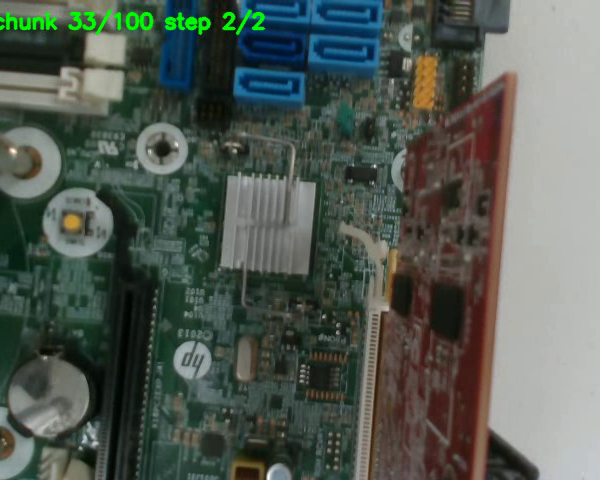}} &
        \adjustbox{valign=m}{\includegraphics[width=0.18\textwidth]{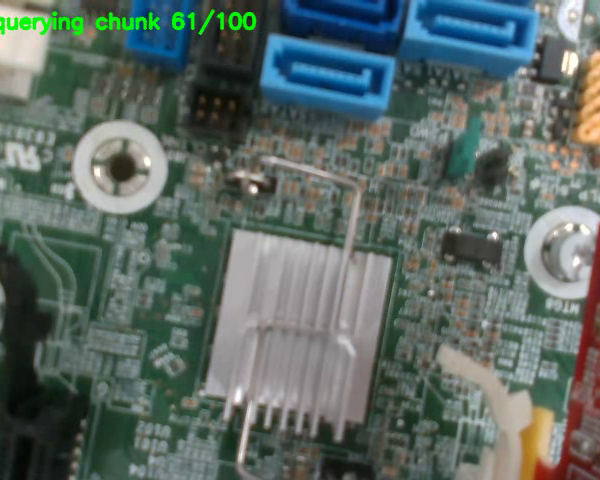}}
    \end{tabular}

    \caption{Representative recovery trajectory under the no-workspace-context
    condition. The top row shows the robot pose and the bottom row shows the
    corresponding wrist-camera observation. Initially, the motherboard is
    completely outside the wrist camera's field of view. As the robot moves,
    the workspace gradually becomes visible, allowing the policy to continue
    toward the target region.}
    \label{fig:no_context_recovery}
\end{figure*}

Figure~\ref{fig:no_context_recovery} illustrates a representative recovery
trajectory under this no-context condition. The top row shows the robot motion,
while the bottom row shows the corresponding wrist-camera observations. At the
beginning of the trajectory, the wrist camera contains no view of the
motherboard. As the robot moves, the workspace gradually enters the camera
field of view, after which the policy continues toward the target region. This
example qualitatively demonstrates recovery from an initial observation that
contains no direct visual evidence of the task workspace.

Table~\ref{tab:initial_view_robustness} reports the quantitative results across
all three initial-view conditions and shows a large gain from 300 to 1,500
generated trajectories in every category. With 1,500 trajectories, the policy
reaches 29/30 and 27/30 precise successes under full and limited context,
respectively. When the workspace is initially absent, it still recovers to the
target neighborhood in 17/30 trials and completes a precise reach in 8/30.
Generated wrist views therefore improve tolerance to initial-view variation,
although performance degrades when direct task evidence is unavailable.

\subsection{Controlled Comparisons and Ablations}
\label{sec:ablations}
\begin{table*}[!t]
\caption{Controlled analysis with $\pi_{0.5}$ (25 trials per target).}
\label{tab:fairness}
\centering
\setlength{\tabcolsep}{8pt}
\begin{tabular}{lccc}
\toprule
Controlled factor & Original setting & Matched setting & Precise success (\%) \\
\midrule
Observation modality & Teleop-150, wrist+side & Teleop-150, wrist only & $58.7 \rightarrow 53.3$ \\
Trajectory timing & GS-900, smoothstep & GS-900, linear & $58.7 \rightarrow 50.7$ \\
\bottomrule
\end{tabular}
\end{table*}

The main comparison intentionally reflects the practical data-collection
setups, but it couples data source with observation modality and trajectory
smoothness.  We therefore conduct two controlled $\pi_{0.5}$ experiments with
25 trials per target.  First, Teleop-150 is retrained with wrist-camera input
only.  Second, the 900-trajectory generated dataset is regenerated with linear
timing in place of the smoothstep function in \eqref{eq:smoothstep}, with every
other setting fixed.  Table~\ref{tab:fairness} shows that both the additional
side view and smooth trajectory supervision contribute to performance, while
the generated wrist-only setting remains directly evaluable.  Data-scale
effects are separately measured by Teleop-50/100/150 and
PrecisionReach-GS-300/900/1500.  Equal episode counts would still not equalize
the quantity or quality of supervision because a teleoperated episode can
contain several corrections, whereas a generated episode follows a clean
structured trajectory.

We further ablate two key pipeline components using 900 generated episodes and
the OpenPI implementation of $\pi_{0.5}$.  Removing the verifier leaves coarse
success unchanged but reduces precise success, consistent with overshoot or
premature termination near the target.  Removing task-keyframe weighting from
the robot-frame alignment degrades both metrics (Table~\ref{tab:ablation_pipeline}).
Start-pose filtering is retained as a validity constraint: disabling workspace,
IK, joint-limit, or manipulability checks produces physically infeasible labels
rather than a meaningful learning baseline.

\begin{table}
\caption{Ablation of key pipeline components with 900 generated episodes.}
\label{tab:ablation_pipeline}
\centering
\setlength{\tabcolsep}{4pt}
\begin{tabular}{lccc}
\toprule
Metric & Full & No verifier & No KF weighting \\
\midrule
Coarse success (\%)  & 93.3 & 93.3 & 74.7 \\
Precise success (\%) & 58.7 & 45.3 & 46.7 \\
\bottomrule
\end{tabular}
\end{table}

\subsection{Reach Verifier Ablation and Robustness}

We further evaluate the reach verifier independently of the VLA policy to
examine its distance-estimation accuracy, architectural components, and
robustness to biased robot state. The evaluation uses 142 held-out episodes
containing 21,476 frames. Each learned variant is trained with three random
seeds, and the best validation checkpoint from each seed is evaluated. During
inference, a stop is triggered after three consecutive predicted distances
below 0.65~cm. A triggered stop is considered correct when the true Cartesian
distance to the instructed target is below 1~cm.

\begin{table*}[t]
    \centering
    \caption{Ablation and robustness analysis of the reach verifier.
    Distance metrics are evaluated using the correct target instruction.
    Near-target MAE is computed over frames whose ground-truth target
    distance is below 3~cm. State robustness is evaluated by applying a
    $-2$~cm bias to the end-effector height.}
    \label{tab:verifier_ablation}
    \setlength{\tabcolsep}{6pt}
    \renewcommand{\arraystretch}{1.12}
    \resizebox{\textwidth}{!}{
    \begin{tabular}{lcccc}
        \hline
        \textbf{Verifier}
        & \textbf{MAE (cm) $\downarrow$}
        & \textbf{Near-Target MAE (cm) $\downarrow$}
        & \textbf{Stop Prec. (\%) $\uparrow$}
        & \textbf{Biased-State Prec. (\%) $\uparrow$} \\
        \hline

        Full
        & $0.64 \pm 0.02$
        & $0.10 \pm 0.02$
        & 99.8
        & $48.1 \pm 11.6$ \\

        w/o spatial attention
        & $0.64 \pm 0.07$
        & $0.12 \pm 0.02$
        & 99.8
        & -- \\

        w/o FiLM
        & $0.67 \pm 0.06$
        & $0.12 \pm 0.03$
        & 99.5
        & -- \\

        w/o instruction conditioning
        & $0.71 \pm 0.05$
        & $0.12 \pm 0.03$
        & 99.3
        & $22.8 \pm 16.1$ \\

        w/o state masking
        & $\mathbf{0.44 \pm 0.01}$
        & $\mathbf{0.08 \pm 0.01}$
        & $\mathbf{100.0}$
        & 0.0 \\

        \hline
    \end{tabular}}
\end{table*}

Table~\ref{tab:verifier_ablation} summarizes the verifier ablations. Under
clean inputs, the full model achieves an overall MAE of 0.64~cm and a
near-target MAE of 0.10~cm, while maintaining 99.8\% episode-level stop
precision. Removing spatial attention has little effect on the overall MAE,
while removing FiLM or instruction conditioning causes modest degradation.
The differences become slightly more apparent near the target, where the full
model achieves 0.10~cm MAE compared with 0.12~cm for the three
language-conditioning ablations.

State masking introduces a clear trade-off between nominal accuracy and
robustness. Without state masking, the clean-input MAE improves from
0.64 to 0.44~cm and the near-target MAE from 0.10 to 0.08~cm. However,
when the end-effector height is biased downward by 2~cm, the same model
fails completely, with episode-level stop precision dropping to 0\%.
In contrast, the full verifier retains $48.1 \pm 11.6$\% stop precision under
the same perturbation. This result indicates that state masking reduces
dependence on the end-effector height, sacrificing some nominal regression
accuracy in exchange for improved robustness to state bias.

Instruction conditioning also contributes to robustness under state
perturbation. Removing instruction conditioning reduces biased-state stop
precision from $48.1 \pm 11.6$\% to $22.8 \pm 16.1$\%, while producing only
a modest change under clean inputs. This suggests that instruction-conditioned
visual features provide an additional cue when the robot-state input becomes
unreliable. In contrast, the small differences between the full model and the
individual FiLM and spatial-attention ablations under clean conditions indicate
that neither mechanism alone dominates the verifier performance.

\subsection{Rendering Fidelity and Robot-Frame Alignment}
\label{sec:representation_diagnostics}

The preceding experiments measure end-to-end behavior. We finally examine two
representation properties that make those rollouts possible: whether rendered
wrist views resemble real observations and whether reconstructed poses are
accurate in the robot frame.

\begin{figure*}[t]
    \centering
    \setlength{\tabcolsep}{1pt}
    \begin{tabular}{@{}c c c c c c@{}}

        \adjustbox{valign=m}{\rotatebox{90}{\textbf{Real}}} &
        \includegraphics[width=0.18\textwidth,valign=m]{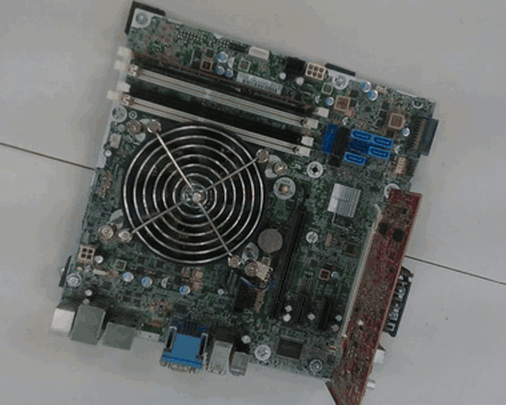} &
        \includegraphics[width=0.18\textwidth,valign=m]{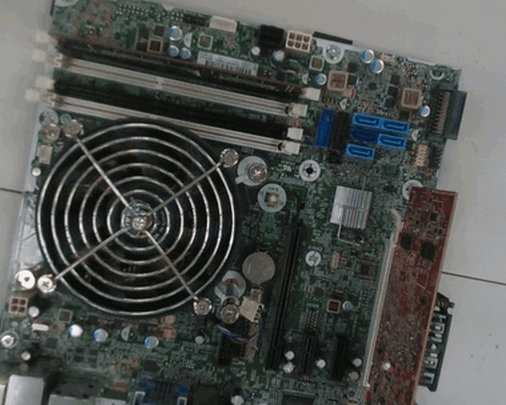} &
        \includegraphics[width=0.18\textwidth,valign=m]{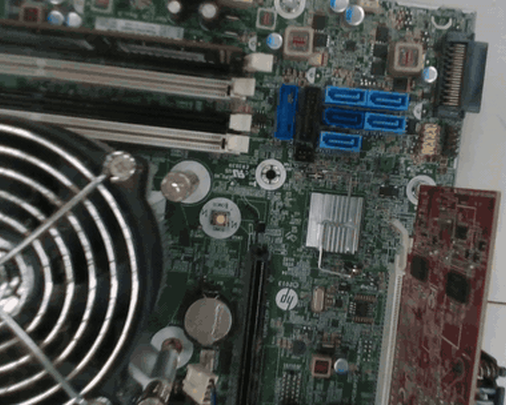} &
        \includegraphics[width=0.18\textwidth,valign=m]{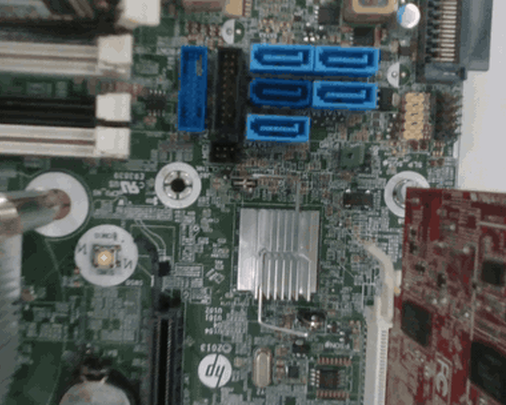} &
        \includegraphics[width=0.18\textwidth,valign=m]{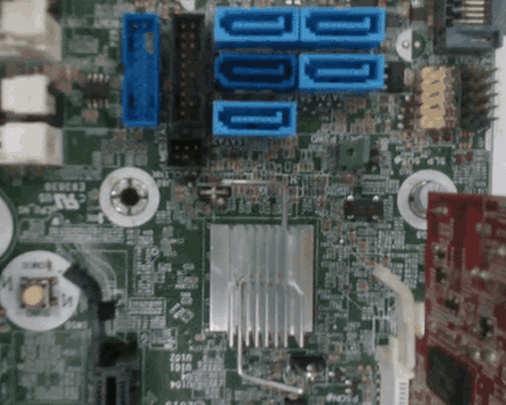} \\
        \noalign{\vskip 3pt}

        \adjustbox{valign=m}{\rotatebox{90}{\textbf{Render}}} &
        \includegraphics[width=0.18\textwidth,valign=m]{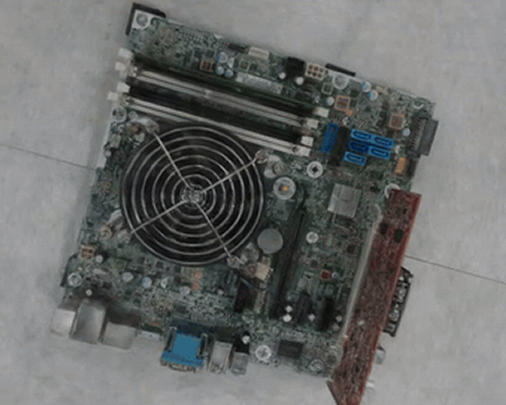} &
        \includegraphics[width=0.18\textwidth,valign=m]{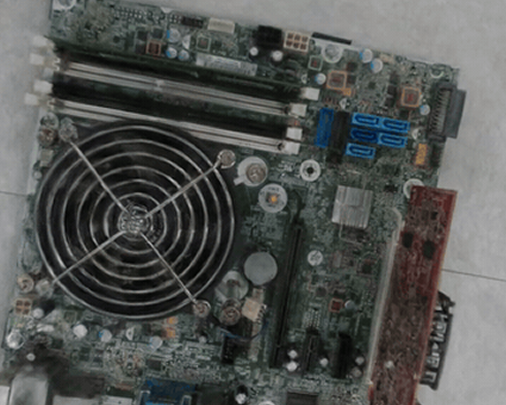} &
        \includegraphics[width=0.18\textwidth,valign=m]{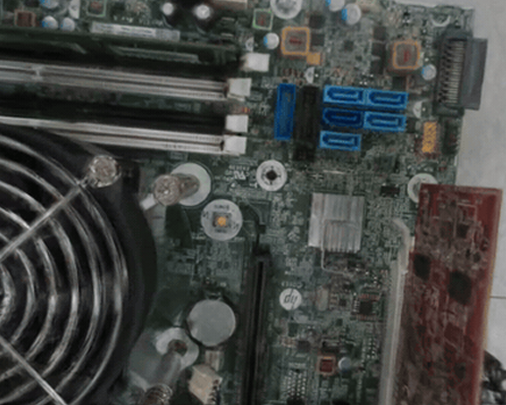} &
        \includegraphics[width=0.18\textwidth,valign=m]{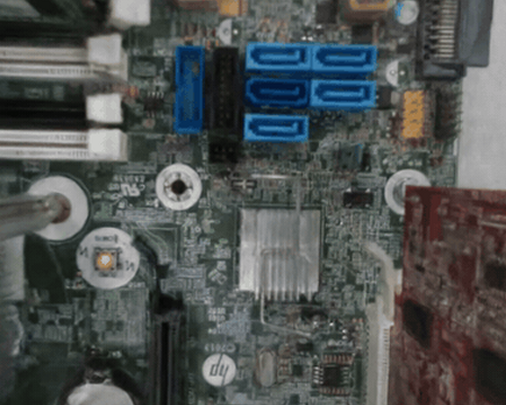} &
        \includegraphics[width=0.18\textwidth,valign=m]{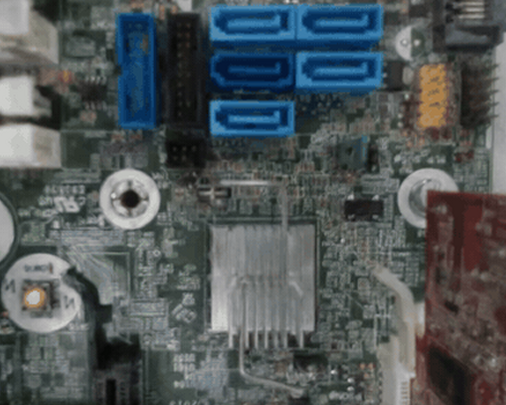}

    \end{tabular}

    \caption{Comparison between real and rendered observations.}
    \label{fig:real_render_comparison}
\end{figure*}

For visual fidelity, we compare real robot videos with videos rendered from the
reconstructed scene along the corresponding camera motion.
Figure~\ref{fig:real_render_comparison} shows representative real and rendered
frames at corresponding viewpoints. The rendered observations closely preserve
the overall scene appearance, component layout, and viewpoint-dependent visual
content of the real wrist-camera observations, while small local differences
remain due to reconstruction and illumination variations.
Table~\ref{tab:visual_similarity} reports complementary image-level metrics.
The SSIM reflects structural and layout consistency, correlation captures brightness and shape
trends, and MSE similarity measures pixel-level agreement. The mean correlation
of 0.843 indicates that the rendered views preserve major appearance trends and
scene layout. The lower SSIM is consistent with the sensitivity of wrist-camera
imagery to small viewpoint and illumination changes and local reconstruction
artifacts.

\begin{table}[!t]
\centering
\caption{Visual similarity between real and rendered robot videos.}
\label{tab:visual_similarity}

\setlength{\tabcolsep}{6pt}
\renewcommand{\arraystretch}{1.18}

\begin{tabular}{lccc}
\hline
Metric & Mean & Best & Worst \\
\hline
SSIM            & 0.682 & 0.919 & 0.503 \\
Correlation     & 0.843 & 0.964 & 0.750 \\
MSE similarity  & 0.979 & 0.991 & 0.963 \\
\hline
\end{tabular}

\renewcommand{\arraystretch}{1.0}
\end{table}

For geometric accuracy, we evaluate the weighted similarity transformation that
maps the COLMAP/3DGS reconstruction into the robot base frame.
Table~\ref{tab:alignment_error} shows centimeter-level position accuracy in
three scenarios, with position RMSE from 0.0094 to 0.0114\,m. The task-keyframe
anchors are tighter, with mean position errors below 0.0040\,m in every
scenario. Mean orientation errors remain below $0.58^\circ$. Together with the
end-to-end results, these measurements show that the representation is
sufficiently aligned for robot-frame wrist-view rendering and feasible reaching
trajectory generation.

\begin{table}[!t]
\centering
\caption{Robot-frame alignment accuracy after transforming the reconstructed scene into the robot coordinate frame.}
\label{tab:alignment_error}
\begin{tabular}{lccc}
\hline
Metric & S1 & S2 & S3 \\
\hline
Position RMSE (m) & 0.0114 & 0.0094 & 0.0096 \\
Position mean error (m) & 0.0102 & 0.0077 & 0.0076 \\
Position max error (m) & 0.0258 & 0.0326 & 0.0329 \\
Anchor position mean error (m) & 0.0039 & 0.0032 & 0.0040 \\
Anchor position max error (m) & 0.0047 & 0.0042 & 0.0067 \\
Orientation mean error (deg) & 0.5792 & 0.4894 & 0.4614 \\
Orientation max error (deg) & 1.1106 & 1.7806 & 1.3140 \\
Anchor orientation mean error (deg) & 0.4573 & 0.2329 & 0.4389 \\
Anchor orientation max error (deg) & 0.5730 & 0.3624 & 0.9064 \\
\hline
\end{tabular}
\end{table}


\section{Discussion and Limitations}
\label{sec:limitations}

Repositioning the motherboard without rebuilding the dataset shows that the policy can tolerate moderate target-location changes rather than memorizing a fixed robot pose. However, this does not imply arbitrary layout generalization; larger workspace or component changes may require a new reconstruction and target demonstrations, after which the automated pipeline can regenerate training trajectories. Performance may also degrade under calibration errors, out-of-distribution viewpoints, persistent occlusion, or severe reconstruction artifacts.

The current system focuses on precise target reaching rather than contact-rich manipulation or post-grasp goal selection. Future work could combine PrecisionReach-GS for long-horizon approach with a contact-aware residual policy using visual, force, or tactile feedback for grasping and subsequent manipulation.

\section{Conclusion}
\label{sec:conclusion}
We presented \textit{PrecisionReach-GS}, a lightweight real-to-sim-to-real
pipeline for data-efficient VLA adaptation in precise robotic reaching.  It
uses a robot-aligned 3DGS workspace to generate wrist-camera observations and
smooth, feasible reaching trajectories without large-scale teleoperation,
object-level meshes, contact models, or a complete physics-based digital twin.
Experiments on motherboard disassembly show reduced active human collection
effort and improved manipulation-ready reaching over limited teleoperation data.
The controlled comparisons and component ablations further identify the roles
of observation modality, smooth trajectory timing, keyframe-weighted alignment,
and stop verification.  These results support robot-aligned visual generation
as a practical route to wrist-camera-only VLA learning in cluttered workspaces.


\bibliographystyle{IEEEtran}
\bibliography{precisionreach_trl}

\end{document}